%% file: template.tex
\documentclass{article}
\usepackage{arxiv}
\usepackage{appendix}
\usepackage{comment}
\usepackage[utf8]{inputenc} 
\usepackage[T1]{fontenc}    
\usepackage[colorlinks=true, linkcolor=blue, urlcolor=blue]{hyperref}       
\usepackage{url}            
\usepackage{booktabs}       
\usepackage{amsfonts}       
\usepackage{nicefrac}       
\usepackage{microtype}      
\usepackage{lipsum}		
\usepackage{graphicx}
\usepackage{hyperref}
\usepackage{xcolor}

\usepackage{doi}
\usepackage{multirow}
\usepackage{multicol}
\usepackage{mdframed} 
\usepackage{xcolor}
\usepackage{subcaption}
\usepackage[normalem]{ulem}
\usepackage{amsmath,amsfonts,bm}
\usepackage{amssymb,amsthm}
\usepackage{enumitem}
\usepackage{subcaption}
\usepackage{adjustbox}
\usepackage{xcolor}  
\usepackage{soul}    
\usepackage{float}
\usepackage[
    backend=biber,
    citestyle=authoryear,
    bibstyle=authortitle,
    mincitenames=1,
    maxcitenames=1,
    maxbibnames=3,
]{biblatex}
\usepackage{svg}
\usepackage{listings}

\definecolor{keywordcolor}{rgb}{0.7, 0.1, 0.1}   
\definecolor{tacticcolor}{rgb}{0.0, 0.1, 0.6}    
\definecolor{commentcolor}{rgb}{0.3, 0.5, 0.3}   
\definecolor{symbolcolor}{rgb}{0.0, 0.1, 0.6}    
\definecolor{sortcolor}{rgb}{0.1, 0.5, 0.1}      
\definecolor{attributecolor}{rgb}{0.7, 0.1, 0.1} 
\definecolor{rulecolor}{rgb}{0, 0, 0}

\newcommand{\citep}[1]{\parencite{#1}}
\setlist[itemize,1]{leftmargin=\dimexpr 18pt}
\setlist[enumerate,1]{leftmargin=\dimexpr 18pt}
\title{
\raisebox{-0.1\height}{\includegraphics[width=0.058\textwidth]{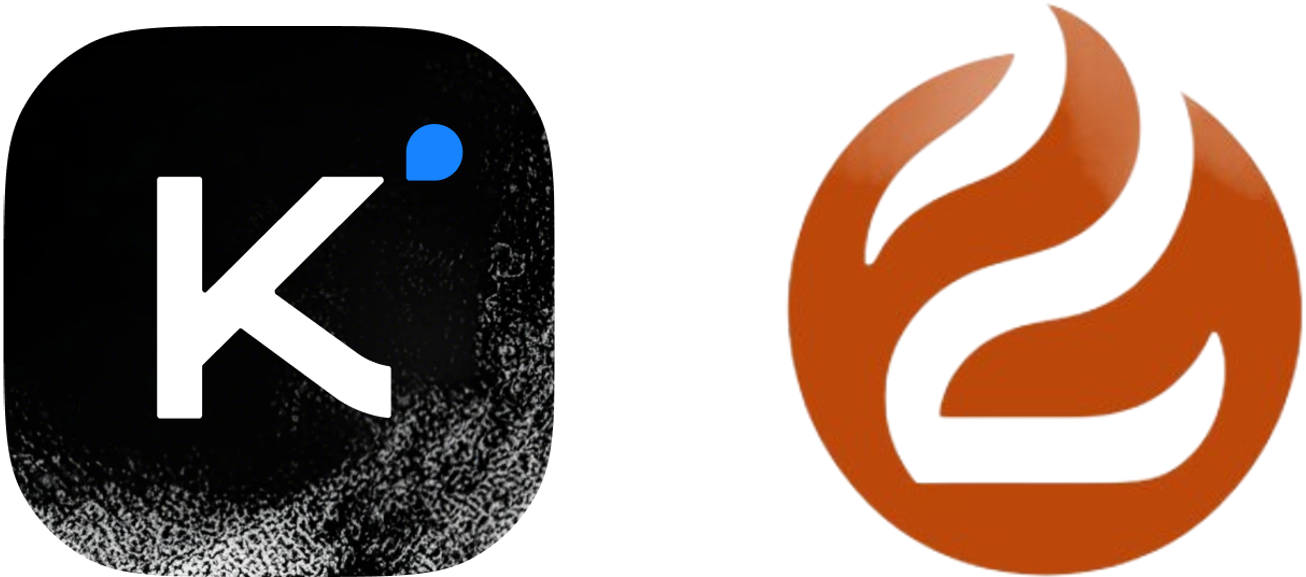}
Kimina-Prover Preview}: Towards Large Formal Reasoning Models with Reinforcement Learning}

\author{Numina \& Kimi Team}

\date{}

\renewcommand{\headeright}{Technical Report}
\renewcommand{\undertitle}{Technical Report of Kimina-Prover Preview}
\renewcommand{\shorttitle}{\raisebox{-0.12\height}{\includegraphics[width=0.02\textwidth]{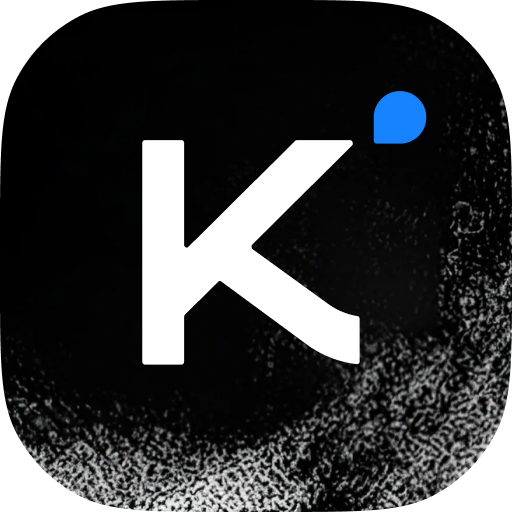} {\includegraphics[width=0.02\textwidth]{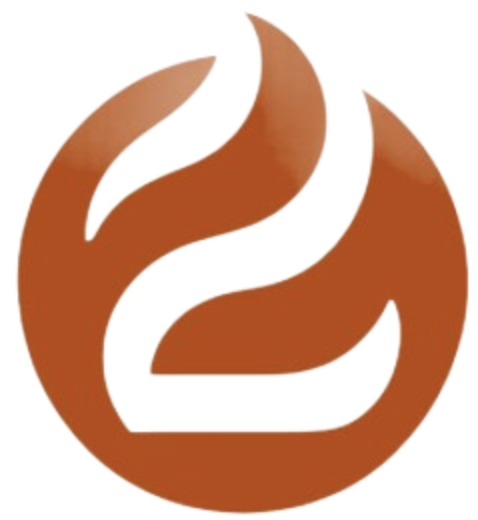}}} 
Kimina-Prover Preview}

\begin{document}
\maketitle

\vspace{-10pt}
\begin{abstract}

We introduce Kimina-Prover Preview, a large language model that pioneers a novel reasoning-driven exploration paradigm for formal theorem proving, as showcased in this preview release. Trained with a large-scale reinforcement learning pipeline from Qwen2.5-72B, Kimina-Prover demonstrates strong performance in Lean 4 proof generation by employing a structured reasoning pattern we term \textit{formal reasoning pattern}. This approach allows the model to emulate human problem-solving strategies in Lean, iteratively generating and refining proof steps. Kimina-Prover sets a new state-of-the-art on the miniF2F benchmark, reaching 80.7\% 
with pass@8192. Beyond improved benchmark performance, our work yields several key insights: (1) Kimina-Prover exhibits high sample efficiency, delivering strong results even with minimal sampling (pass@1) and scaling effectively with computational budget, stemming from its unique reasoning pattern and RL training; (2) we demonstrate clear performance scaling with model size, a trend previously unobserved for neural theorem provers in formal mathematics; (3) the learned reasoning style, distinct from traditional search algorithms, shows potential to bridge the gap between formal verification and informal mathematical intuition. We open source distilled versions with 1.5B and 7B parameters of Kimina-Prover\footnote{\url{https://github.com/MoonshotAI/Kimina-Prover-Preview}}.

\end{abstract}

\begin{figure}[H]
    \centering
    \includegraphics[scale=0.4]{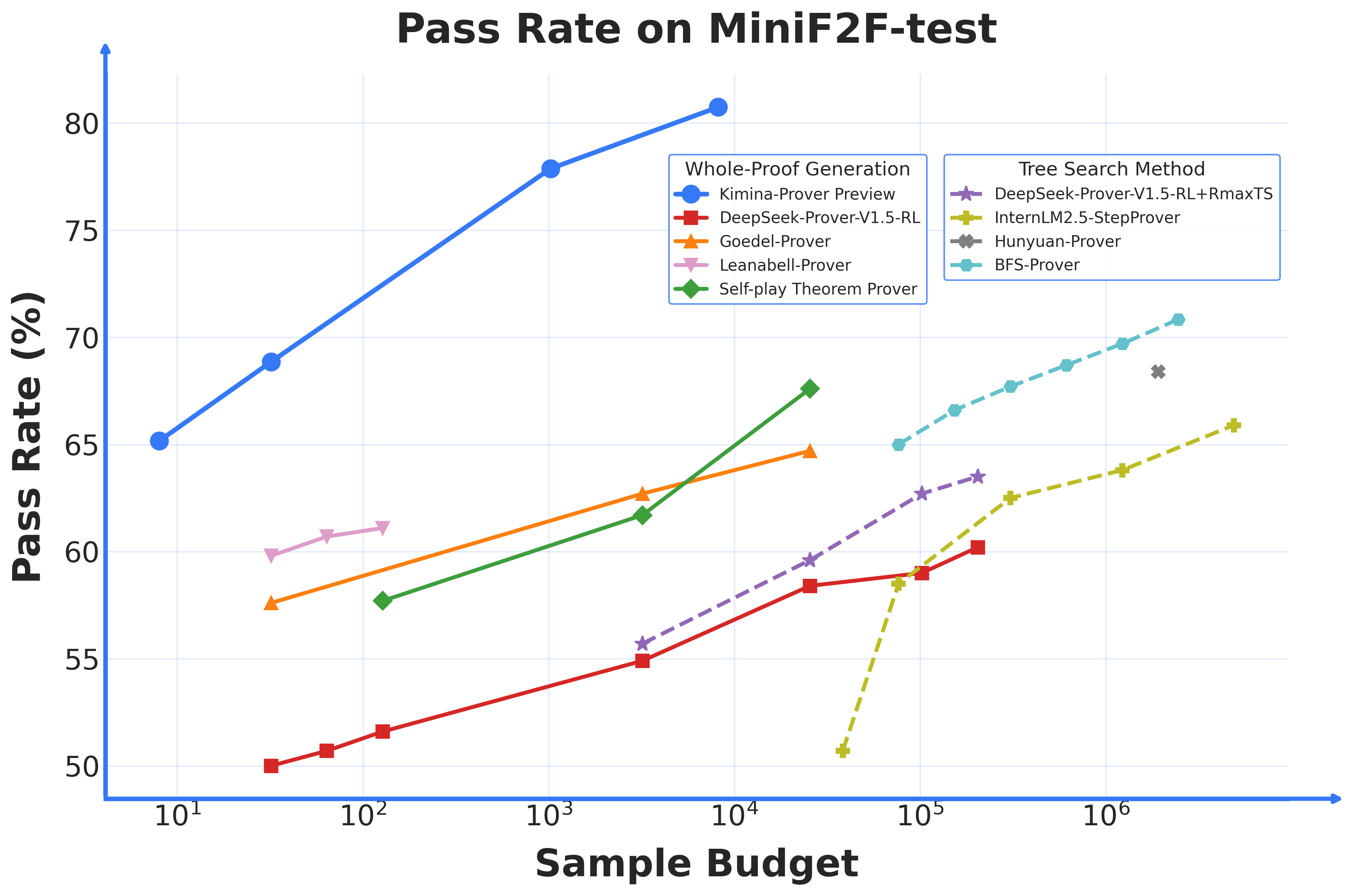}
    \caption{Performance comparison of different theorem proving methods on miniF2F-test dataset. The $x$-axis shows the sample budget (number of language model queries), and the $y$-axis shows the pass rate. Methods are divided into two categories: whole-proof generation and tree search approaches. Results demonstrate that Kimina-Prover Preview achieves the highest pass rate with fewer samples,  while tree search methods generally require more samples to achieve comparable performance.}
    \label{fig:main_result}
\end{figure}

\section{Introduction}

Recent advances in neural theorem proving have focused on leveraging large language models (LLMs) to tackle the inherent challenges of formal reasoning in proof assistants like Lean 4 \citep{lean4paper} or Isabelle \citep{nipkow2002isabelle}. Initial approaches focused on training LLMs to generate individual proof steps or tactics within interactive proof assistants \citep{polu2022formalmathematicsstatementcurriculum, wu2024internlm25stepproveradvancingautomatedtheorem, wang2023dt-solver, deepmind2024alphaproof}, often coupling the LLM's predictive capabilities with classical search algorithms like Best-First Search (BFS) \citep{polu2022formalmathematicsstatementcurriculum, wu2024internlm25stepproveradvancingautomatedtheorem, xin2025bfsproverscalablebestfirsttree} or Monte Carlo Tree Search (MCTS) \citep{NEURIPS2022_a8901c5e, wang2023dt-solver, deepmind2024alphaproof} to explore the proof space. Other strategies involved training models to generate entire proof structures from a given state \citep{xin2024deepseekproverv15harnessingproofassistant, lin2025goedelproverfrontiermodelopensource, first2023baldurwholeproofgenerationrepair}. Despite notable progress, these existing methods face significant challenges. The reliance on explicit search algorithms (BFS, MCTS) introduces substantial computational overhead and complexity, limiting scalability. Furthermore, while LLMs excel at pattern matching and sequence generation, effectively capturing the deep, structured, and often non-linear reasoning required for complex formal proofs remains difficult. Standard supervised fine-tuning or basic chain-of-thought prompting may not sufficiently elicit the sophisticated reasoning necessary. Critically, previous neural theorem provers tailored for formal mathematics have generally not demonstrated clear improvements in performance corresponding to increases in model size, suggesting limitations in their ability to leverage larger model capacity for enhanced reasoning. While some works \citep{lin2025leanstar, wang2025malotmultiagentleanbasedlong} attempted to integrate informal reasoning hints, they typically relied on shorter chain-of-thought patterns or models not specifically optimized for long-form reasoning via reinforcement learning.

In this work, we introduce Kimina-Prover Preview, representing an early attempt to bridge this gap by pioneering a novel reasoning-driven exploration paradigm for formal theorem proving. Built upon the Kimi k1.5 reinforcement learning (RL) pipeline, which has demonstrated success in eliciting long chain-of-thought reasoning for complex informal math and coding tasks \citep{team2025kimi}, Kimina-Prover is specifically adapted for formal reasoning within the Lean 4 proof assistant. Instead of relying on external search algorithms, our approach leverages the LLM's own internal reasoning capabilities, enhanced through large-scale RL with carefully designed reward signals and structured reasoning patterns. This allows the model to emulate human problem-solving strategies, implicitly exploring the proof space and iteratively generating and refining proof steps through a process guided by its internal reasoning tokens.
Our contributions include:

\begin{itemize}
    \item \textbf{Pioneering Reasoning-Driven Exploration.} We pioneer the application of large-scale reinforcement learning to elicit long chain-of-thought reasoning in theorem proving.
    \item \textbf{Effective Model Scaling for Formal Mathematics.} We show clear improvements in neural theorem proving performance as LLM size increases, a scaling effect not observed in previous systems.
    \item \textbf{State-of-the-Art Performance.} Illustrated in Figure~\ref{fig:main_result}, Kimina-Prover achieves state-of-the-art performance by reaching 80.7\% with pass@8192 on the miniF2F benchmark, significantly surpassing prior SotA achieved by BFS Prover (\cite{xin2025bfsproverscalablebestfirsttree}) (72.95\%).
\end{itemize}

\section{Methodology}

\begin{figure}[t]
    \centering
    \begin{subfigure}[b]{0.2\textwidth}
        \centering
        \includegraphics[width=\linewidth]{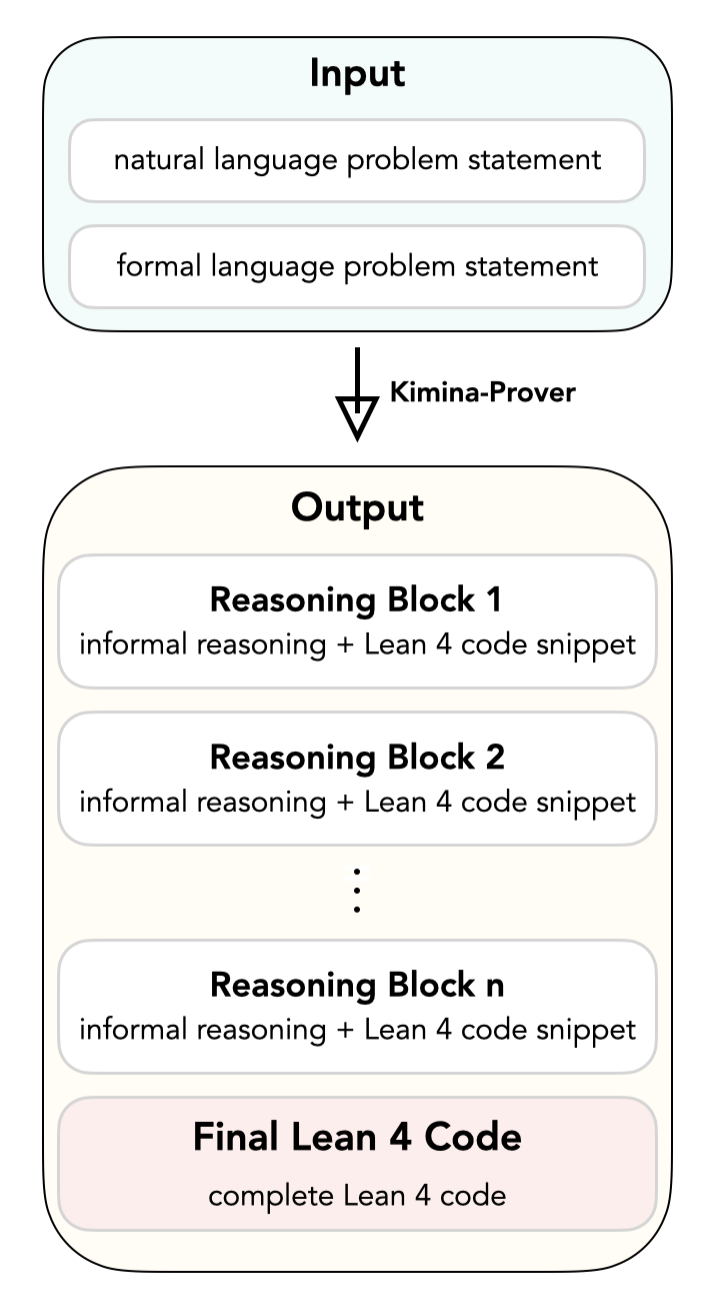}
    \end{subfigure}
    \hfill
    \begin{subfigure}[b]{0.75\textwidth}
        \centering
        \includegraphics[width=\linewidth]{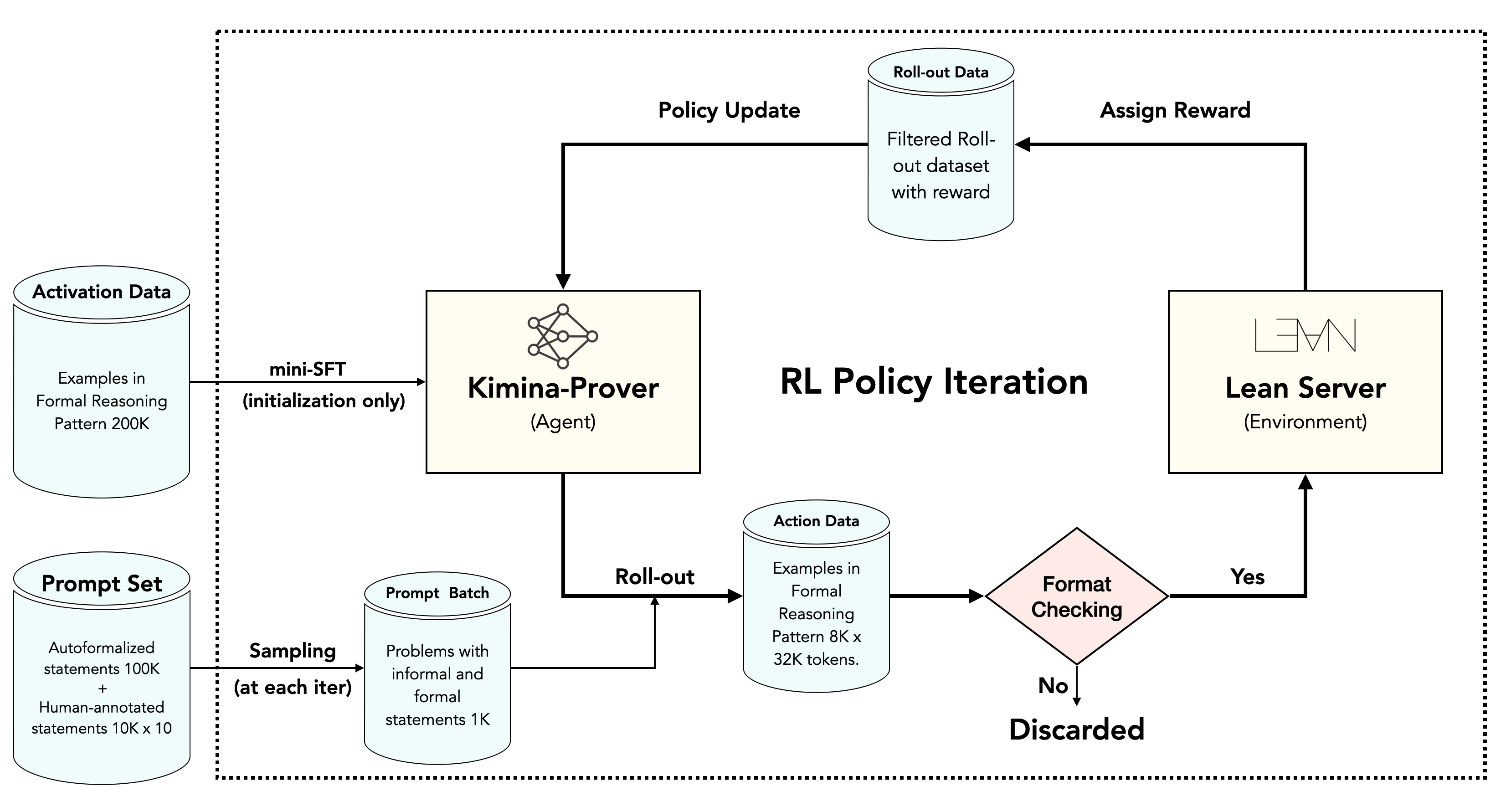}
    \end{subfigure}
    \caption{\textbf{Left.} Formal Reasoning Pattern. \textbf{Right.} Formal RL pipeline.}
    \label{fig:main}
    \vspace{-3mm}
\end{figure}

\subsection{Autoformalization for Constructing a Diverse Base Problem Set}

To enable online reinforcement learning for formal theorem proving, we require a large, diverse set of formal problems in Lean 4. Manual construction of such a dataset is costly and time-intensive. To address this, we train models to automatically translate natural language problem statements into syntactically valid Lean 4 code ending with a placeholder proof. However, this task presents a fundamental challenge: the lack of a concrete, automatic reward signal. Unlike proof search, where success is easily defined by whether a theorem is proven, there is no straightforward way to verify the correctness of a generated formal problem statement with respect to the natural language input. Our solution combines careful initialization, supervised fine-tuning, and a structured expert iteration process with LLM-based judging to progressively improve quality. We open source our model \href{https://huggingface.co/AI-MO/Kimina-Autoformalizer-7B}{Kimina-Autoformalizer-7B} for the community and detail our training recipe in Appendix \ref{training_autoformalizer}.

\subsection{Formal Reasoning Pattern}

One of the critical challenges for reasoning LLMs to excel at formal theorem proving is the lack of alignment between informal mathematical reasoning data and its translation to formal proofs. To tackle this, we design a novel formal reasoning pattern to enable Kimina-Prover to \textit{think} in an environment that aligns the informal and formal mathematical reasoning. During training, we filter for responses that format their \textit{thinking} in between \texttt{<think> \dots 
\: </think>} tokens and output the final proof between chosen special tokens. Within the \textit{thinking} block, we seek informal-formal alignment by interspersing informal reasoning with relevant Lean 4 code snippets, also marked by special tokens. To enforce \textit{thinking} block alignment with the final proof, we ensure that the majority of the Lean 4 code snippets appear in the final proof. With this reasoning pattern (see Figure~\ref{fig:main} on the left), we observe scaling of output token length, which correlates with passing proof-complexity from our evaluations.

Additionally, this pattern also offers great improvement in terms of reasoning explainability compared to search-based provers. The \textit{thinking} block allows users to inspect the models' internal process during proof generation. This provides tangible insights about the nature of the failure modes of the model and serves as an educational tool for end-users.

\textbf{Cold start.} To kickstart the models' ability to output formal proofs following our formal reasoning pattern, we perform a minimal supervised fine-tuning run before starting our reinforcement learning training loop. In particular, we collect a dataset of olympiad-style mathematics problems with statements and solutions both in natural language format and formalized in Lean 4. We then ask Claude 3.7 Sonnet~\citep{anthropic2025claude} to synthesize a \textit{thinking} block output combining the the informal and formal proofs to synthesize a mini-SFT dataset of around 20K examples (we have tried several LLMs and only Claude's performance is satisfying). This significantly boosts the models' ability to align the informal and formal reasoning steps and enhances downstream performance during the RL phase.

\textbf{Informal math mix-training.} To further bolster the initial informal mathematical capabilities of our model, we incorporate informal mathematical thinking data from Kimi k1.5 into the SFT training. The synthetic data generated by Claude exhibited a limited range of reflection patterns. While we observed the emergence of test-time scaling and reflection during subsequent RL training, these reflections were often of low quality, characterized by repetition and meaningless phrases. Integrating informal thinking data aims to provide a better starting point for generating more meaningful reflections. Indeed, our experiments during the RL phase demonstrate that checkpoints trained with this mixed data significantly outperform those trained solely on formal reasoning data.

\subsection{Reinforcement Learning}

Following the supervised fine-tuning (SFT) phase, we employ reinforcement learning (RL) to further enhance our model's formal theorem-proving capabilities, as illustrated in the pipeline schematic (Figure \ref{fig:main}). The RL process iteratively refines the model's policy. Each iteration begins by sampling a batch of $N=1000$ problems from our established problem set. For each problem, the current policy generates $k=8$ candidate solutions (rollouts). The final Lean 4 code of each candidate is then rigorously verified using the Lean compiler to determine correctness. A binary reward signal is assigned: $1$ for a completely correct proof and $0$ otherwise. In line with the methodology of Kimi k1.5~\citep{team2025kimi}, 
we employ the following loss to optimize the language model:

\begin{equation}
    L(\theta) = \mathbb{E}_{(x,y^*) \sim D} \left[ \mathbb{E}_{(y,z) \sim \pi_{\text{old}}} \left[ r(x, y, y^*) - \tau \log Z - \tau \log \frac{\pi_\theta(y, z | x)}{\pi_{\text{old}}(y, z | x)} \right] \right]\, ,
    \label{eq:loss}
\end{equation}

where $\pi_{\text{old}}$ is the previous policy model and the normalization constant $\log Z $ is approximated using the empirical mean of the rewards in practice. 
During RL training, we see strong scaling of formal reasoning at test time. However, limited SFT data and formal structure cause early RL format collapse from negative gradients. To prevent this, we apply format filtering with two key constraints: (1) each generated sample must contain at least one tactic block; and (2) tactic blocks must collectively cover at least 60\% of the Lean code included in the final Lean 4 solution. Moreover, To counter format collapse from negative gradients, we randomly discard samples with negative gradients (probability $\omega=0.5$). Combined with mixed training on informal math data, this stabilizes training and promotes more sophisticated formal reasoning.

In practice, our RL training is conducted starting from Qwen2.5-72B~\citep{qwen2.5}. We maintain a constant learning rate of $(2 \times 10^{-6})$ and a fixed KL divergence coefficient $\tau=0.4$ (Equation \ref{eq:loss}). This KL constraint ensures stability by controlling the policy's deviation from the initial supervised policy throughout the RL process.

\section{Results}

\subsection{Inference Setup}

\textbf{Benchmark.} We evaluate our models on the miniF2F benchmark \citep{zheng2022minif2fcrosssystembenchmarkformal}, using the Lean 4 version from DeepSeek-ProverV1.5. To prevent data contamination, we perform 13-gram decontamination and explicitly remove all AMC12, AIME, and IMO problems from the Numina Math 1.5 training set if their sources overlap with problems in the miniF2F test set. We also identify and correct eight unsolvable problems in the benchmark (\lstinline{mathd_numbertheory_618}, \lstinline{aime_1994_p3}, \lstinline{amc12a_2021_p9}, \lstinline{mathd_algebra_342}, \lstinline{mathd_numbertheory_343}, \lstinline{mathd_algebra_158}, \lstinline{induction_pord1p1on2powklt5on2}, \lstinline{induction_prod1p1onk3le3m1onn}), releasing the corrected versions via the Numina HuggingFace repository. Evaluations utilize a 32K token context length and sampling budgets up to $8192$ attempts, with each attempt sampled independently.

\textbf{Distillation.} We train our 1.5B and 7B models by rolling out data from the Kimina-Prover-Preview model with 72B parameters and performing SFT, initializing from Qwen2.5-Math-1.5B-Instruct and Qwen2.5-Math-7B-Instruct, respectively. We use packing and cosine learning rate scheduling with $lr = 2 \times 10^{-5}$ for 3 epochs.

\textbf{Lean Server.}
In our reinforcement learning and evaluation pipelines, we integrate the Numina Lean Server \citep{numinaleanserver} as the verification backend to provide real-time feedback for generated proof attempts. Built upon Lean FRO's LeanREPL \citep{leanrepl}, the Numina Lean Server employs an LRU-based caching mechanism that reuses preloaded environments based on \texttt{import} headers, significantly reducing initialization overhead. Furthermore, it supports extensive parallelization across multiple CPUs by managing multiple Lean REPL processes concurrently. These innovations result in a $10\times$ speedup in verification throughput, achieving up to 100 iterations per second on machines equipped with 64 CPU cores and 512 GB RAM. During RL training, this verification system operates efficiently in the rollout phase, evaluating proofs in real time as they are generated. Due to the relatively slow proof-generation process, verification does not become a bottleneck, requiring only 640 CPU cores for training. This efficiency contrasts with previous language model-based theorem proving approaches, which typically require thousands of CPU cores to sustain real-time verification at scale \citep{NEURIPS2022_a8901c5e, xin2024deepseekproverv15harnessingproofassistant}.

\subsection{Performance Analysis}

\subsubsection{Comparison with State-of-the-Art Methods} 
On the miniF2F benchmark, Kimina-Prover Preview achieves a state-of-the-art result among all evaluated systems—including both whole-proof generation and tree search methods—reaching 80.74\% miniF2F-test accuracy (see Table~\ref{tab:minif2f}). Importantly, Kimina-Prover demonstrates strong performance even in low-pass settings and scales effectively with higher sampling budgets. With just pass@1, the model already achieves 52.94\%, and with pass@32, it reaches 68.85\%, already competitive with or surpassing many larger-sample baselines, showcasing exceptional sample efficiency. This efficiency can be attributed to Kimina-Prover's novel reasoning process. Rather than relying on explicit step-based searches, Kimina-Prover implicitly flattens the step-wise search, allowing the model to decide both how and what to search. This architectural flexibility enables more targeted and dynamic reasoning, leading to higher performance with fewer samples.

\input{tables/minif2f_results}

\begin{figure}
    \centering
    \begin{minipage}[t]{0.45\textwidth}
        \centering
        \includegraphics[width=\linewidth]{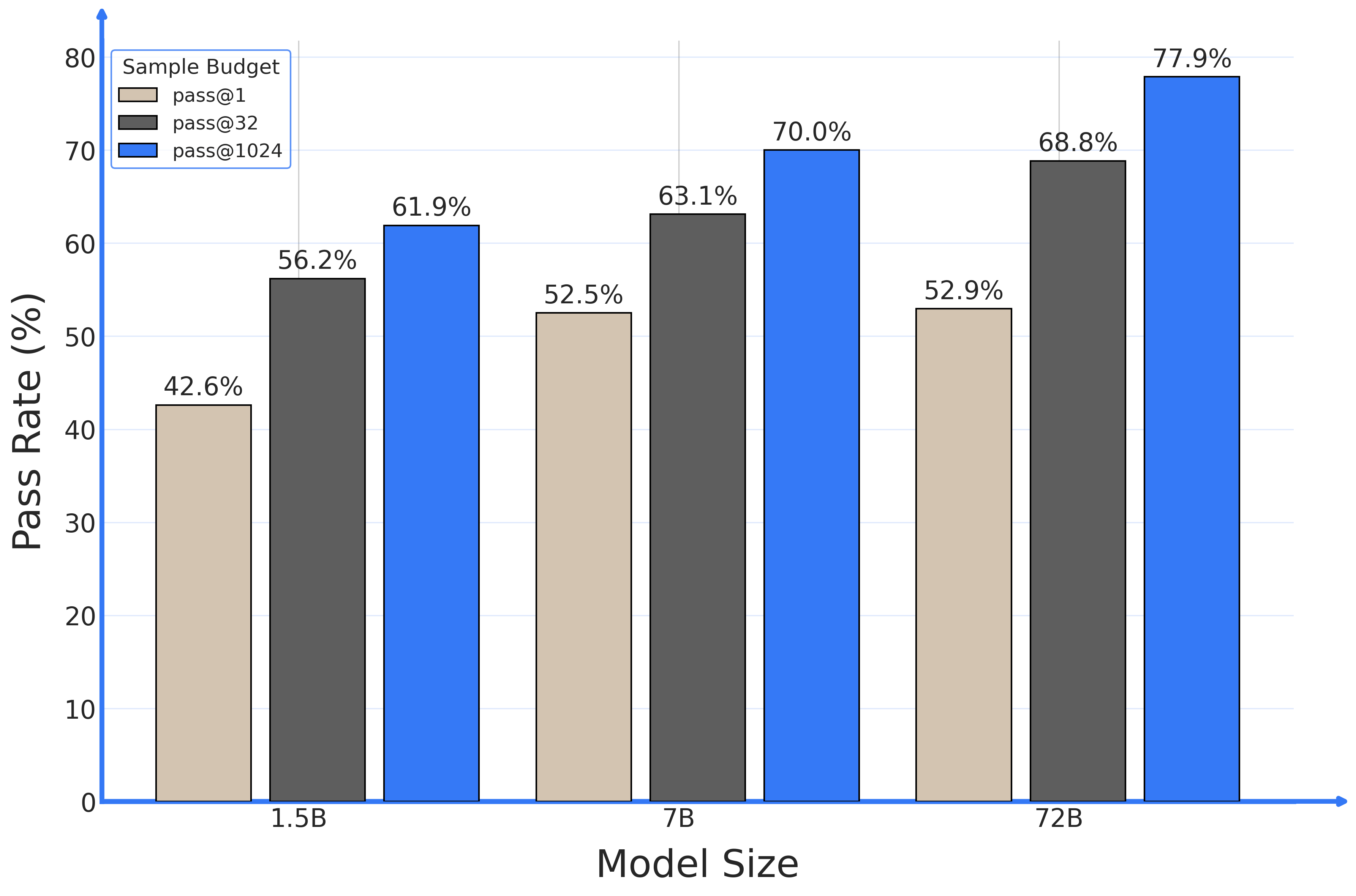}
        \caption{Performance scaling of Kimina-Prover \\ models across different sizes. }\label{fig:size-scaling}
    \end{minipage}
    \hfill
    \begin{minipage}[t]{0.45\textwidth}
        \centering
        \includegraphics[width=\linewidth]{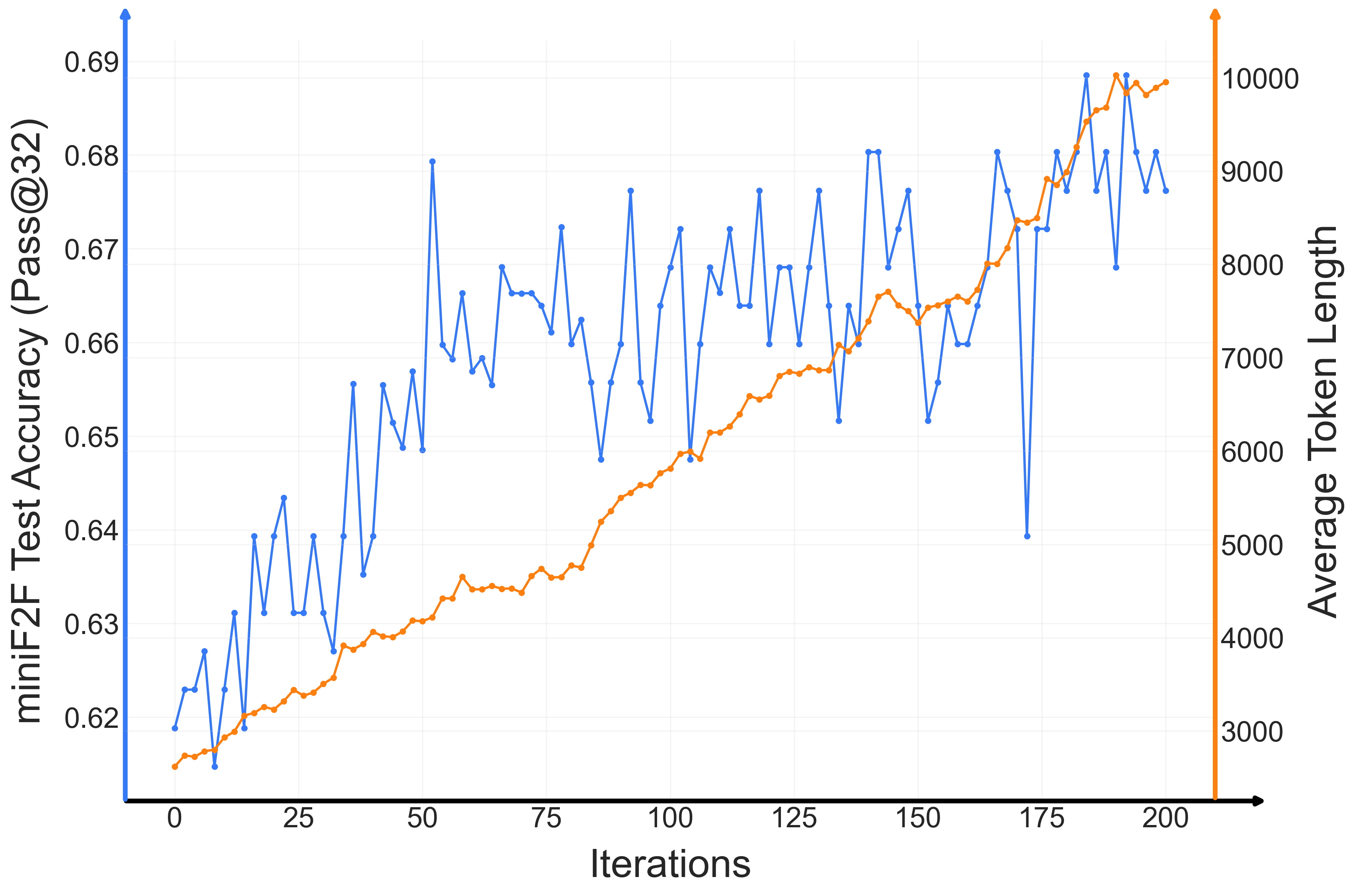}
        \caption{miniF2F accuracy (pass@32) and average output token length of Kimina-Prover during training.}
        \label{fig:test-time-scaling}
    \end{minipage}
    \vspace{-5mm}
\end{figure}

Another key observation (see Figure~\ref{fig:size-scaling}) is that our models present a clear upward trend in performance as model size increases - from 1.5B, to 7B, and finally 72B. Especially with larger sampling budgets, the 72B variant shows significant gains (the 72B model outperforms the 7B version by +0.44\%, +5.75\%, and +7.87\% at those respective sampling budgets.). To our knowledge, this is the first formal reasoning system that consistently scales in performance with model size, suggesting that Kimina-Prover not only scales computationally but also in its reasoning capabilities.

\subsubsection{Comparison with General Purpose LLMs} 

In Table~\ref{tab:aime_imo}. we compare Kimina-Prover with leading general-purpose reasoning models—OpenAI’s o3 and Gemini 2.5 Pro—on the miniF2F benchmark and its IMO and AIME subsets. Despite these subsets containing relatively easier, older problems, Kimina-Prover significantly outperforms both models across the board. At pass@32, Kimina-Prover achieves 68.85\% on miniF2F, versus 37.70\% for Gemini and 24.59\% for o3. On the IMO and AIME subsets, Kimina-Prover scores 20.00\% and 46.67\%, respectively—well above Gemini (5\%, 13.33\%) and o3 (0\%, 6.67\%). At pass@8192, Kimina-Prover further improves to 80.74\%, 40.00\% (IMO), and 86.67\% (AIME). These results highlight a core limitation of general-purpose models. o3 and similar systems fail at formal reasoning, defaulting to informal, unverifiable answers. Gemini shows formal reasoning capabilities but often suffers from hallucination and generates invalid proofs. In contrast, Kimina-Prover consistently generates formally verifiable, Lean-checkable proofs, demonstrating both accuracy and reasoning ability.

\input{tables/aime_imo}

\subsubsection{Gap between Informal Mathematics and Formal Mathematics} 

One particularly interesting finding is that our formal reasoning model shows strong potential to bridge the gap between formal and informal mathematics. As shown in Table~\ref{tab:aime_imo}, while general-purpose reasoning models such as Gemini-2.5-pro and o3-mini are capable of solving all AIME problems in miniF2F under informal settings, they exhibit substantially lower performance in formal mathematics. This discrepancy suggests that transferring domain knowledge from general mathematical problem solving to formal mathematical reasoning remains a challenging task. In contrast, Kimina-Prover’s formal reasoning capabilities demonstrate that formal mathematics can complement and enhance informal reasoning, rather than exist in isolation. By learning to reason within a formal system, the model appears to gain deeper structural understanding, which could in turn benefit informal mathematical problem solving. This opens exciting future directions, where formal mathematics may not only be useful for verification, but also for boosting model performance on informal math reasoning tasks.

\subsubsection{Test Time Scaling in Formal Reasoning} 
A key capability of reasoning models is their ability to improve with increased test-time budget. Figure~\ref{fig:test-time-scaling} shows how Kimina-Prover’s miniF2F pass@32 accuracy (blue) correlates with the average token length of its outputs (orange) over training. As the model learns to generate longer proofs—from ~2,500 to over 10,000 tokens—its accuracy rises from ~61.8\% to nearly 69\%. Unlike informal math models, which often scale smoothly, formal reasoning shows a much more volatile pattern. Accuracy jumps are frequent and sometimes regress, especially in the mid-phase between 50–150 iterations, likely due to the model adjusting to complex, structured reasoning with limited formal training data. This turbulent but ultimately successful trend highlights how formal reasoning can still scale effectively, even without vast pretraining corpora. This success suggests that similar reasoning approaches may be transferable to other domains that are also limited in domain knowledge during pre-training.

\subsubsection{Emergent human-like proof style.}
Given the described formal reasoning pattern, we observe that after initialization and RL training, our model demonstrates the ability to produce complex reasoning patterns. These include exploring multiple reasoning paths, reflecting on and refining its thinking process, and analyzing small-scale cases to uncover general patterns (see Appendix \ref{appendix:thinking_output_analysis}). Additionally, the proofs generated by Kimina-Prover are more decomposed and structured compared to those of previous step-based provers. This is exhibited by the abundance of \lstinline{have} statements within the proofs, which is a common pattern observed in human-written proofs optimized for clarity (see Appendix \ref{appendix:proof_comparison}). These emergent behaviors of reflection, decomposition, and refinement scale effectively with increasing problem difficulty, giving our approach a distinct advantage over models that extensively rely on existing Lean 4 automation tools.

\section{Conclusion}

In conclusion, we present Kimina-Prover, a large reasoning model for Lean 4 theorem proving developed through a training recipe combining autoformalization, SFT, and RL with a specific formal reasoning pattern. Our key findings demonstrate that performance scales significantly with both context length and model size—a trend not typically observed in tree-search provers—leading to state-of-the-art results (80.74\% on miniF2F pass@8192) with modest compute. This underscores the potential of reasoning-enabled neural provers. Promising future directions include enhancing proof quality by filtering outputs that overuse high-level tactics, enabling iterative refinement using Lean compiler feedback to fix errors efficiently, and integrating external tools like library search and computation engines to alleviate generation challenges.

\medskip
\printbibliography

\newpage

\appendix

\section{Contributions}

\begin{multicols}{2}
\noindent
\textbf{Numina} \\
Jia Li* \\
Mert Unsal* \\
Mantas Baksys* \\
Marco Dos Santos* \\
Marina Vinyes* \\
Zhenzhe Ying* \\
Zekai Zhu* \\
Jianqiao Lu* \\
Hugues de Saxcé* \\
Bolton Bailey \\
Ebony Zhang \\
Frederick Pu \\
Jiawei Liu \\
Jonas Bayer \\
Julien Michel\\
Léo Dreyfus-Schmidt \\
Lewis Tunstall \\
Luigi Pagani \\
Moreira Machado \\
Pauline Bourigault \\
Ran Wang \\
Stanislas Polu \\
Thibaut Barroyer\\
Wen-Ding Li \\
Yazhe Niu \\
Yann Fleureau \\
Zhouliang Yu\\

\columnbreak
\textbf{Kimi Team} \\
Haiming Wang* \\
Zhengying Liu* \\
Xiaohan Lin* \\
Junqi Liu* \\
Flood Sung* \\
Chendong Song \\
Chenjun Xiao \\
Dehao Zhang \\
Han Zhu \\
Longhui Yu \\
Yangyang Hu  \\
Zhilin Yang \\
Zihan Wang \\

\end{multicols}

Names marked with an asterisk (*) indicate core contributors. Other authors are listed in alphabetical order based on their first names.

\section{Related Work}
Recent approaches to automated theorem proving have combined large language models with formal proof assistants such as Lean 4. These systems typically employ a language model trained to either generate individual proof steps \citep{NEURIPS2022_a8901c5e, polu2022formalmathematicsstatementcurriculum,deepmind2024alphaproof,wang2023dt-solver,wu2024internlm25stepproveradvancingautomatedtheorem, xin2025bfsproverscalablebestfirsttree, lin2025goedelproverfrontiermodelopensource,wang2024provingtheoremsrecursively} or produce entire proof completions from a given proof state \citep{xin2024deepseekproverv15harnessingproofassistant, lin2025goedelproverfrontiermodelopensource,first2023baldurwholeproofgenerationrepair,huang2024mustardmasteringuniformsynthesis,wang2023legoproverneuraltheoremproving,jiang2023draftsketchproveguiding}.

To enhance exploration in proof search, these language-model-based approaches are frequently integrated with classical tree search algorithms, such as Best-First Search \citep{polu2022formalmathematicsstatementcurriculum, wu2024internlm25stepproveradvancingautomatedtheorem, xin2025bfsproverscalablebestfirsttree} and Monte Carlo Tree Search \citep{NEURIPS2022_a8901c5e, wang2023dt-solver, deepmind2024alphaproof}. The search component plays a central role in these systems, as it guides the exploration and selection of promising proof paths based on heuristic evaluations. Such hybrid methods significantly benefit from heuristic-driven exploration but incur substantial computational overhead.

Large-scale reinforcement learning has recently been applied to improve the reasoning abilities of language models, exemplified by models such as OpenAI's o1 \citep{openai2024openaio1card}, DeepSeek's R1 \citep{deepseekai2025deepseekr1incentivizingreasoningcapability}, and Kimi's k1.5 \citep{team2025kimi}. These models, trained with extensive RL on carefully engineered reward signals, exhibit emergent long chain-of-thought reasoning behavior, enabling them to achieve impressive results in complex mathematical and coding tasks, including competitions like AIME and Codeforces. These achievements underscore the potential for long, structured reasoning in overcoming challenging reasoning problems.

A few recent works have attempted to integrate informal reasoning with formal theorem proving. For instance, \cite{lin2025leanstar} proposed generating informal thoughts prior to predicting individual tactics, and \cite{wang2025malotmultiagentleanbasedlong} employs a structured interaction between informal natural language reasoning and formal verification in Lean 4. However, these previous attempts either relied on conventional short-form chain-of-thought reasoning or relied on reasoning models trained via transfer learning from general reasoning tasks and supervised fine-tuning rather than reinforcement learning. Prior to our work, the feasibility and effectiveness of applying RL-driven long-form reasoning directly within formal theorem proving remained unexplored.

\section{Training Details of Kimina-Prover}

\subsection{Informal Dataset}
The informal dataset serves as the foundational layer of our pipeline, feeding directly into supervised fine-tuning (SFT) and reinforcement learning (RL) for the large formal reasoning model. Below, we outline the data processing steps that convert the raw informal mathematical dataset into a curated prompt set for training our models.

Our informal data pipeline consists of the following steps:

\begin{enumerate}
    \item \textbf{Initial Dataset Acquisition:} We begin with the Numina 1.5 dataset \citep{li2024numinamath}, a comprehensive collection of mathematical problems.

    \item \textbf{Filtering and Preprocessing:} We filter this dataset based on specific criteria, retaining only problems clearly classified as either proofs or problems with explicit numeric or symbolic outputs. Problems involving geometry and combinatorics are excluded to form a dataset more suitable for autoformalization. This filtered dataset is denoted by \textit{auto-statement-candidates}.

    \item \textbf{Autoformalization:} The \textit{auto-statement-candidates} dataset undergoes an automatic formalization process, resulting in the \textit{auto-statements} dataset. This step converts natural language mathematical problems into formal statements compatible with the Lean 4 proof assistant.

    \item \textbf{Human Annotation and Refinement:} To ensure the quality and precision of formalizations, we established a dedicated annotation team tasked with reviewing and refining the outputs from the autoformalization process. Annotated outputs are categorized as follows:
        \begin{itemize}
            \item \textit{Human Statements:} Statements refined by human annotators.
            \item \textit{Human Proofs:} Select, challenging statements are further annotated with formal proofs by domain experts. These proofs constitute a significant portion of our supervised fine-tuning (SFT) dataset.
        \end{itemize}

    \item \textbf{Prompt Set Creation:} The refined formalized statements from the \textit{auto-statements} and \textit{human statements} datasets are combined to form our final prompt set. The validated and proven formal statements subsequently enrich the SFT dataset, creating a dynamic cycle of continuous improvement for subsequent iterations of reinforcement learning.
\end{enumerate}

This structured and iterative approach ensures robust data quality, facilitating effective model training and iterative refinement in subsequent phases.

\subsection{Training Details of Kimina-Autoformalizer}
\label{training_autoformalizer}

\subsubsection{Model Initialization}

We begin by creating a supervised fine-tuning dataset for autoformalization, to teach the model the structure and style of competition-level Lean 4 problems. This dataset aggregates formal problem pairs from multiple sources: \textbf{PutnamBench}, \textbf{miniF2F}, \textbf{ProofNet}, and \textbf{Compfiles} \citep{compfiles2024} \citep{zheng2022minif2fcrosssystembenchmarkformal} \citep{tsoukalas2024putnambenchevaluatingneuraltheoremprovers} \citep{azerbayev2023proofnetautoformalizingformallyproving}. We ablate the contribution of each source and find that all positively contribute to downstream performance. In particular, including Mathlib data, such as the MMA dataset \citep{jiang2023multilingualmathematicalautoformalization} - an LLM-generated informalization of Mathlib - degrades performance. We hypothesize two key reasons: (1) the mathematical content in Mathlib is substantially different in nature and tone from competition problems, and (2) Mathlib statements often involve auxiliary variables and rely on external definitions, which may confuse the model during generation. Our initialization is a fine-tuned version of Qwen2.5-Coder-7B-Instruct on the curated dataset, capable of producing syntactically valid Lean 4 problem statements from informal descriptions. 

\subsubsection{Expert Iteration with LLM Judge}

After supervised initialization, we employ a structured expert iteration loop to improve the quality and coverage of the autoformalization model. This loop uses a challenging subset of competition-style problems from the NuminaMath 1.5 dataset. For each iteration, we begin by sampling a minibatch of informal problems. For each problem, the model generates a number of candidate formalizations. These are filtered to retain only those that compile successfully in Lean 4. We then employ a QwQ-32B as a judge to evaluate the semantic correctness of each remaining formalization, using a handcrafted prompt to guide its assessment. We observe that using multiple samples from the judge model and unanimous voting structure significantly reduces the amount of false positives with little impact on true positives. Formalizations that pass both the compilation and judging stages are added to the training dataset, and a training step is performed. 

Since the reward signal in this task is inherently fuzzy, and LLM-based feedback is susceptible to false positives, we monitor model outputs throughout training with Lean 4 experts and gradually improve the judge prompt over iterations. This allows us to maintain fine-grained control over the data quality. As our proving infrastructure improves, we also introduce automated filters to discard problematic formalizations. These include detecting logical contradictions, proving the negation of the formalized statement, or identifying that the problem is trivial via a short LLM-generated proof. These safeguards help ensure the training set remains both challenging and correct, guiding the model toward meaningful improvements in future iterations.

\subsubsection{Kimina-Autoformalizer Performance}

\begin{figure}[h!]
    \centering
    \includegraphics[width=0.5\linewidth]{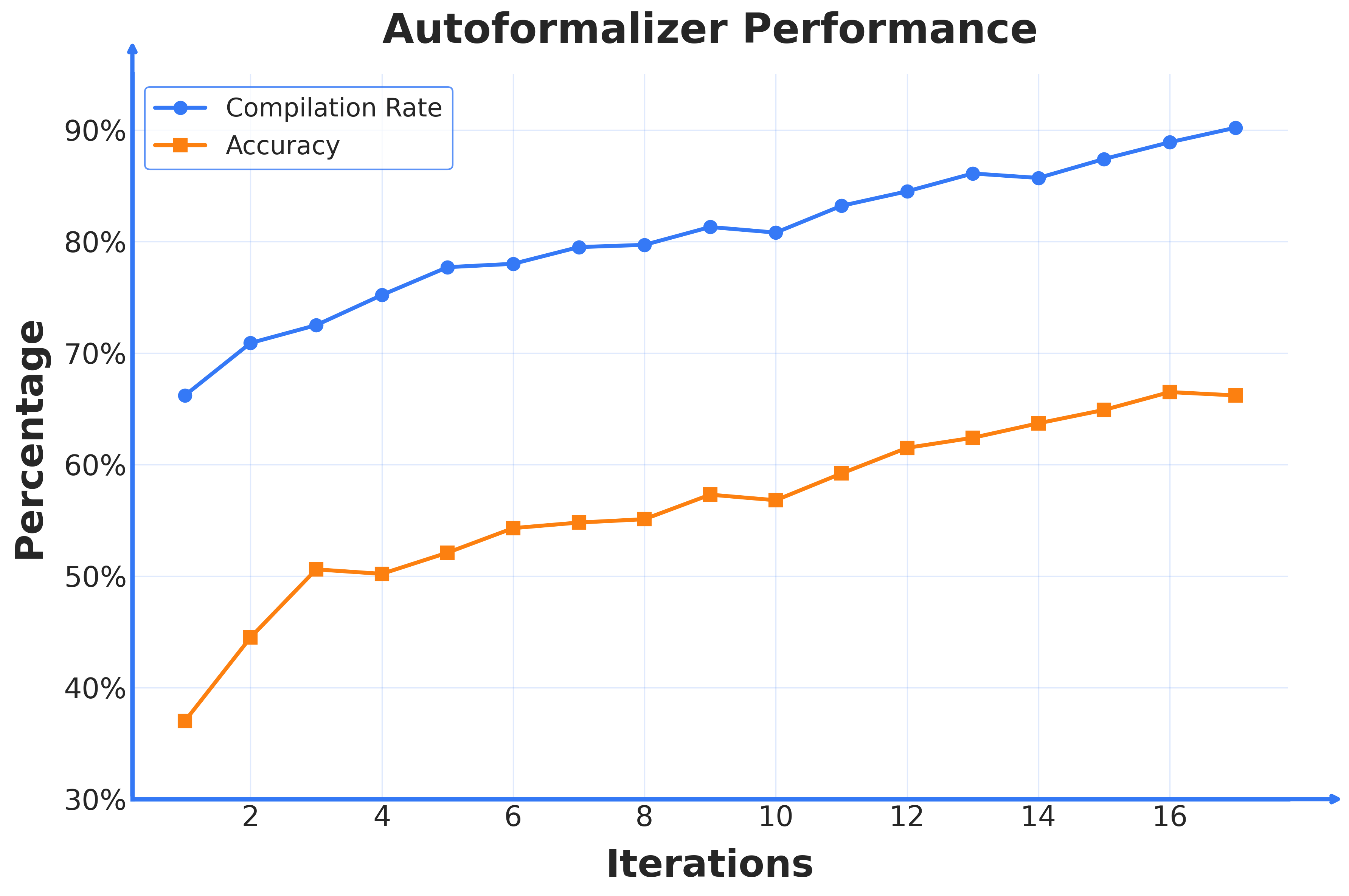}
    \caption{Performance of Autoformalizer Across Iterations.}
    \label{fig:enter-label}
\end{figure}

To evaluate our autoformalizer, we use a human-curated test set of size approximately 1,000 and optimize an LLM judge prompt on this set for reliable results. After each iteration, we generate one autoformalization per problem using greedy decoding. We track two metrics: (1) Lean 4 compilation rate and (2) autoformalization accuracy, defined as the percentage of samples that both compile and are judged correct by the LLM. We observe steady and reliable improvement over iteration as can be seen in the plots, reaching 90\% one-shot compilation rate and 66\% accuracy.

We evaluate several existing models for autoformalization tasks, but found that most struggled to generate valid Lean 4 code consistently. Instead, these models predominantly produced Lean 3 syntax, a limitation we attribute to the composition of their pretraining data. Prior to these LLMs' knowledge cutoff dates, Lean 3 dominated online repositories and documentation, while Lean 4—with its substantial syntax differences—remained relatively new and underrepresented. This training data imbalance caused models without specific Lean 4 fine-tuning to default to the more familiar Lean 3 patterns. The inability of comparison models to produce compilable Lean 4 code at meaningful rates ultimately prevented us from establishing fair benchmarks against existing approaches, highlighting the challenges of working with emerging formal languages.

We note that this method required careful expert monitoring as we observe that the model would repeatedly make mistakes that the LLM judge model cannot catch, as long as it passed compilation. We believe this presents a significant challenge in applying reinforcement learning in non-verifiable domains.

\subsection{Problem Set}

In this section, we detail the creation of our problem set, which serves as a foundational component of our reinforcement learning pipeline. The prompt set consists of two distinct subsets: one derived from the autoformalization model and another consisting of human-annotated statements. To ensure balanced difficulty across the problem set, we utilize the QwQ-32B preview model to assign a difficulty rating to each problem and subsequently construct a dataset with an evenly distributed difficulty spectrum.

Given that our annotated statements are limited in number—approximately 10k—compared to the 100k generated by autoformalization, we prioritize high-quality data in training. Therefore, we resample our annotated set to achieve a 1:1 ratio with the autoformalized subset, yielding a balanced and high-quality problem set of 200k total problems. 

To further improve the efficiency and effectiveness of the reinforcement learning pipeline, we iteratively refine the prompt set using the following strategies:

\textbf{Error Filtering via Negation Proving}. We employ DeepSeek-Prover's negation-proving to identify and remove potentially erroneous formalizations from our problem set.

\textbf{Adaptive Problem Pruning}. After specific iterations of reinforcement learning, we prune problems where the model consistently demonstrates high proficiency. This adaptive pruning allows the model to concentrate computational resources on increasingly challenging and informative problems, promoting continuous improvement in formal reasoning capabilities.

\textbf{Annotation Pipeline for Difficult Problems}. For wrongly formalized statements or very difficult problems, we send them to the annotation pipeline for additional statement and proof annotation by human annotators. This ensures that challenging problems are not discarded but receive appropriate human oversight.
 
\textbf{Post-RL Validation}. After reinforcement learning training, we use a judge model to assess whether the proofs generated by the model are correct or if the model has merely leveraged a mistake in the formalization. This step ensures the integrity and reliability of the generated proofs.

\section{Illustrative Proof Examples found by Kimina-Prover}

\subsection{Newly Proved IMO problem from the miniF2F Benchmark}

Kimina-Prover successfully proves an IMO problem (\lstinline{imo_1968_p5_1}) that no previous publicly released model has ever solved. We present the proof generated by Kimina-Prover below. This proof shows an excellent ability of the model to \textbf{plan} its proof sketch. Such emergent behavior in the output shows a clear improvement in reasoning ability compared to previous models. For example, in this proof, the model begins by first stating a key auxiliary lemma \lstinline{h2} and continues to prove it across multiple lines. This is followed by guessing the correct answer via the (\lstinline{use 2 * a}) tactic. The proof is completed via a complex reasoning chain, where the previously proved lemma \lstinline{h2} is key and necessary to complete the proof.

\begin{lstlisting}[frame = single, framesep=3mm, framerule=0.5mm, caption = {Lean 4 proof of IMO 1968 P5 ($1^{st}$ part) found by Kimina-Prover.}]
theorem imo_1968_p5_1 (a : ℝ) (f : ℝ → ℝ) (h₀ : 0 < a)
    (h₁ : ∀ x, f (x + a) = 1 / 2 + Real.sqrt (f x - f x ^ 2)) : ∃ b > 0, ∀ x, f (x + b) = f x := by
  have h2 : ∀ x, 0 ≤ f x ∧ f x ≤ 1 := by
    intro x
    have h1 := h₁ (x - a)
    rw [show x - a + a = x by ring] at h1
    have h3 : 0 ≤ Real.sqrt (f (x - a) - f (x - a) ^ 2) := Real.sqrt_nonneg (f (x - a) - f (x - a) ^ 2)
    have h4 : f x = 1 / 2 + Real.sqrt (f (x - a) - f (x - a) ^ 2) := by linarith
    have h5 : f x ≤ 1 := by
      have h6 : Real.sqrt (f (x - a) - f (x - a) ^ 2) ≤ 1 / 2 := by
        have h7 : f (x - a) - f (x - a) ^ 2 ≤ 1 / 4 := by
          nlinarith [sq_nonneg (f (x - a) - 1 / 2)]
        have h8 : 0 ≤ Real.sqrt (f (x - a) - f (x - a) ^ 2) := Real.sqrt_nonneg (f (x - a) - f (x - a) ^ 2)
        have h9 : Real.sqrt (f (x - a) - f (x - a) ^ 2) ≤ Real.sqrt (1 / 4 : ℝ) := Real.sqrt_le_sqrt (by linarith)
        have h10 : Real.sqrt (1 / 4 : ℝ) = 1 / 2 := by
          rw [Real.sqrt_eq_iff_sq_eq] <;> norm_num
        linarith
      linarith [h4, h6]
    have h11 : 0 ≤ f x := by
      nlinarith [Real.sqrt_nonneg (f (x - a) - f (x - a) ^ 2), h4]
    exact ⟨h11, h5⟩
  use 2 * a
  constructor
  · linarith [h₀]
  · intro x
    have h3 := h₁ (x + a)
    have h4 := h₁ x
    rw [show x + a + a = x + 2 * a by ring] at h3
    have h5 : f (x + 2 * a) = 1 / 2 + Real.sqrt (f (x + a) - f (x + a) ^ 2) := by linarith
    have h6 : f (x + a) - f (x + a) ^ 2 = (f x - 1 / 2) ^ 2 := by
      have h7 : f (x + a) = 1 / 2 + Real.sqrt (f x - f x ^ 2) := by linarith
      rw [h7]
      ring_nf
      rw [Real.sq_sqrt (by nlinarith [h2 x])]
      ring
    rw [h6] at h5
    have h7 : Real.sqrt ((f x - 1 / 2) ^ 2) = abs (f x - 1 / 2) := by
      rw [Real.sqrt_sq_eq_abs]
    have h8 : abs (f x - 1 / 2) = f x - 1 / 2 := by
      have h9 : f x ≥ 1 / 2 := by
        have h10 := h₁ (x - a)
        rw [show x - a + a = x by ring] at h10
        have h11 : Real.sqrt (f (x - a) - f (x - a) ^ 2) ≥ 0 := Real.sqrt_nonneg (f (x - a) - f (x - a) ^ 2)
        linarith [h10, h11]
      apply abs_of_nonneg
      linarith
    rw [h7, h8] at h5
    linarith [h5]
\end{lstlisting}

\subsection{Example Proofs of AIME Problems}

Kimina-Prover excels at solving complex trigonometric equations in the formal theorem proving domain. This is shown in its proof of \lstinline{aime_1991_p9} presented below. Given the initial hypothesis \lstinline{h0} stated as: \lstinline{h0 : 1 / Real.cos x + Real.tan x = 22 / 7}, the model is able to use the trigonometric identity 

\begin{lstlisting}[frame = single, framesep=3mm, framerule=0.5mm]
have h3 : Real.sin x ^ 2 + Real.cos x ^ 2 = 1 := Real.sin_sq_add_cos_sq x 
\end{lstlisting}

to derive the values for the sine and cosine functions.

\begin{lstlisting}[frame = single, framesep=3mm, framerule=0.5mm]
have h_cos : Real.cos x = 308 / 533 := by nlinarith [h_cos_sq]
have h_sin' : Real.sin x = 435 / 533 := by
    have h9 : Real.sin x = (22 * Real.cos x - 7) / 7 := h_sin
    rw [h9, h_cos]
    norm_num
\end{lstlisting}

From here, the model substitutes these derived values into \lstinline{h1}: \lstinline{h1 : 1 / Real.sin x + 1 / Real.tan x = m}

After this step, it successfully closes out the proof showing great mastery of algebraic techniques.

\begin{lstlisting}[frame = single, framesep=3mm, framerule=0.5mm, caption = {Lean 4 proof of AIME 1991 Problem 9 found by \texttt{Kimina-Prover}.}]
theorem aime_1991_p9 (x : ℝ) (m : ℚ) (h0 : 1 / Real.cos x + Real.tan x = 22 / 7)
    (h1 : 1 / Real.sin x + 1 / Real.tan x = m) : ↑m.den + m.num = 44 := by
  have h2 : Real.cos x ≠ 0 := by
    by_contra h
    have h3 : 1 / Real.cos x = 0 := by
      field_simp [h]
    have h4 : Real.tan x = 0 := by
      rw [Real.tan_eq_sin_div_cos]
      simp [h]
    have h5 : (22 : ℝ) / 7 = 0 := by linarith [h₀, h3, h4]
    norm_num at h5

  have h_tan : Real.tan x = Real.sin x / Real.cos x := by
    rw [Real.tan_eq_sin_div_cos]

  have h3 : Real.sin x ^ 2 + Real.cos x ^ 2 = 1 := Real.sin_sq_add_cos_sq x

  have h4 : 1 / Real.cos x + Real.tan x = 22 / 7 := h₀
  rw [h_tan] at h4
  have h5 : 1 / Real.cos x + Real.sin x / Real.cos x = 22 / 7 := h4
  have h6 : (1 + Real.sin x) / Real.cos x = 22 / 7 := by
    have h7 : Real.cos x ≠ 0 := h2
    field_simp [h7] at h5 ⊢
    nlinarith

  have h7 : 7 * (1 + Real.sin x) = 22 * Real.cos x := by
    have h8 : Real.cos x ≠ 0 := h2
    field_simp at h6
    nlinarith

  have h_sin : Real.sin x = (22 * Real.cos x - 7) / 7 := by linarith

  have h8 : Real.sin x ^ 2 + Real.cos x ^ 2 = 1 := Real.sin_sq_add_cos_sq x
  have h9 : Real.sin x = (22 * Real.cos x - 7) / 7 := h_sin

  have h_cos_sq : Real.cos x ^ 2 = (308 / 533) ^ 2 := by
    have h10 : Real.cos x ≠ 0 := h2
    have h11 : Real.sin x = (22 * Real.cos x - 7) / 7 := h_sin
    rw [h11] at h3
    have h12 : ((22 * Real.cos x - 7) / 7) ^ 2 + Real.cos x ^ 2 = 1 := by linarith
    have h13 : (22 * Real.cos x - 7) ^ 2 + (7 * Real.cos x) ^ 2 = 49 := by
      nlinarith
    have h14 : (Real.cos x) * (533 * Real.cos x - 308) = 0 := by
      ring_nf at h13 ⊢
      linarith
    cases' (mul_eq_zero.mp h14) with h_cos h15
    · -- Real.cos x = 0
      exfalso
      exact h10 (by linarith)
    · -- 533 * Real.cos x - 308 = 0
      have h_cos : Real.cos x = 308 / 533 := by linarith
      nlinarith [h_cos]
  have h_cos : Real.cos x = 308 / 533 := by nlinarith [h_cos_sq]

  have h_sin' : Real.sin x = 435 / 533 := by
    have h9 : Real.sin x = (22 * Real.cos x - 7) / 7 := h_sin
    rw [h9, h_cos]
    norm_num

  have h_csc_cot : (1 / Real.sin x : ℝ) + (1 / Real.tan x : ℝ) = (29 / 15 : ℝ) := by
    have h10 : Real.sin x ≠ 0 := by
      by_contra h
      have h11 : Real.sin x = 0 := by linarith
      rw [h11] at h3
      have h12 : Real.cos x ^ 2 = 1 := by nlinarith
      have h13 : Real.cos x = 1 ∨ Real.cos x = -1 := by
        have h14 : Real.cos x ^ 2 - 1 = 0 := by linarith
        have h15 : (Real.cos x - 1) * (Real.cos x + 1) = 0 := by
          ring_nf at h14 ⊢
          linarith
        cases (mul_eq_zero.mp h15) with
        | inl h16 => left; linarith
        | inr h17 => right; linarith
      cases h13 with
      | inl h14 =>
        have h15 : Real.cos x = 308 / 533 := h_cos
        linarith
      | inr h14 =>
        have h15 : Real.cos x = 308 / 533 := h_cos
        linarith
    have h11 : Real.tan x ≠ 0 := by
      by_contra h
      have h12 : Real.sin x = 0 := by
        rw [Real.tan_eq_sin_div_cos] at h
        have h13 : Real.cos x ≠ 0 := h2
        field_simp [h13] at h
        linarith
      contradiction
    have h12 : (1 / Real.sin x : ℝ) + (1 / Real.tan x : ℝ) = (29 / 15 : ℝ) := by
      have h13 : Real.tan x = Real.sin x / Real.cos x := Real.tan_eq_sin_div_cos x
      rw [h13]
      have h14 : Real.sin x ≠ 0 := h10
      have h15 : Real.cos x ≠ 0 := h2
      field_simp [h14, h15, h_sin', h_cos]
      norm_num
    exact h12
  have h_m : (m : ℝ) = (29 / 15 : ℝ) := by
    have h12 : (1 / Real.sin x : ℝ) + (1 / Real.tan x : ℝ) = (29 / 15 : ℝ) := h_csc_cot
    have h13 : (1 / Real.sin x : ℝ) + (1 / Real.tan x : ℝ) = (m : ℝ) := by
      exact_mod_cast h₁
    rw [h13] at h12
    linarith
  have h14 : m = (29 / 15 : ℚ) := by
    have h15 : (m : ℝ) = (29 / 15 : ℝ) := h_m
    have h16 : (m : ℝ) = (29 / 15 : ℚ) := by
      norm_num at h15 ⊢
      exact_mod_cast h15
    exact_mod_cast h16
  rw [h14]
  norm_num    
\end{lstlisting}

Kimina-Prover also excels at carrying out long calculations using the Lean 4 tactic \lstinline{calc}. This is shown in the proof of \lstinline{aime_1997_p9} presented below. We would especially like to draw attention to the calc blocks required to prove statements \lstinline{h32, h33, ..., h34}. in this proof. These calculation steps require careful and precise consideration of the current proof context and cannot rely solely on Lean 4's automation tactics, which shows deep understanding of the formal theorem proving domain.

\newpage

\subsection{Example of Formal Reasoning Pattern}
The following list demonstrates the formal reasoning pattern of Kimina-Prover. Upon receiving the problem, the model first uses natural language reasoning to get a general proof idea. Then, it proceeds to formalize its thoughts to Lean 4 code step by step. In this process, the model further ensures the rigor of the reasoning and deliberates on how to effectively utilize appropriate tactics to formalize the proof. Upon completion of the proof, the model assembles all Lean code snippets to form a coherent and complete Lean proof.
\begin{mdframed}[linewidth=0.5mm]
Input:
\begin{lstlisting}[frame = single]
-- In an arithmetic sequence, the 7th term is 30, and the 11th term is 60. Determine the 21st term.
theorem mathd_algebra_354 (a d : ℝ) (h₀ : a + 6 * d = 30) (h₁ : a + 10 * d = 60) :
    a + 20 * d = 135 := by
\end{lstlisting}
\end{mdframed}

\begin{mdframed}[linewidth=0.5mm]
Output:

<think>

\# Solving for the 21st Term of an Arithmetic Sequence

I'm given that in an arithmetic sequence, the 7th term is 30, and the 11th term is 60. I need to find the 21st term.

First, let's think about the structure of an arithmetic sequence. The general term of an arithmetic sequence is given by $a_n = a + (n-1)d$, where $a$ is the first term and $d$ is the common difference.

Given:

- The 7th term $a_7 = a + 6d = 30$

- The 11th term $a_{11} = a + 10d = 60$

- We need to find the 21st term $a_{21} = a + 20d$

Let's solve for $a$ and $d$ to find $a_{21}$.

Subtracting the first equation from the second gives:

$(a + 10d) - (a + 6d) = 60 - 30$

$4d = 30$

$d = \frac{30}{4} = 7.5$ or $\frac{15}{2}$

Now, substituting $d = 7.5$ back into the first equation:

$a + 6(7.5) = 30$

$a + 45 = 30$

$a = 30 - 45 = -15$

So, the 21st term is $a_{21} = a + 20d = -15 + 20(7.5) = -15 + 150 = 135$

Let's formalize this in Lean 4:

\begin{lstlisting}[frame = single]
import Mathlib
import Aesop

set_option maxHeartbeats 0

open BigOperators Real Nat Topology Rat

theorem mathd_algebra_354 (a d : ℝ) (h₀ : a + 6 * d = 30) (h₁ : a + 10 * d = 60) :
    a + 20 * d = 135 := by
\end{lstlisting}

We have two equations:

1. $a + 6d = 30$ ($h_0$)

2. $a + 10d = 60$ ($h_1$)

We need to find $a + 20d$.

From the two equations, we can eliminate $a$ to solve for $d$. Subtracting $h_0$ from $h_1$:

$(a + 10d) - (a + 6d) = 60 - 30$

$4d = 30$

$d = 30/4 = 7.5$ or $\frac{15}{2}$

Let's compute this in Lean:

\begin{lstlisting}[frame = single]
have hd : d = 15 / 2 := by
  linarith
\end{lstlisting}

Now we have $d = \frac{15}{2}$. Let's verify this calculation:

From $a + 10d = 60$ and $a + 6d = 30$, subtracting gives $4d = 30$, so $d = \frac{30}{4} = \frac{15}{2}$. This is correct.

Next, we'll substitute $d = \frac{15}{2}$ back into one of the equations to solve for $a$. Let's use $h_0$: $a + 6d = 30$

$a + 6(\frac{15}{2}) = 30$

$a + 45 = 30$

$a = 30 - 45 = -15$

In Lean, once we have $d$, we can determine $a$:

\begin{lstlisting}[frame = single]
have ha : a = -15 := by
  linarith [h₀, hd]
\end{lstlisting}

Now we need to compute $a + 20d$:
$a + 20d = -15 + 20(\frac{15}{2}) = -15 + 150 = 135$

In Lean, we can directly compute $a + 20d$ using the values of $a$ and $d$:

\begin{lstlisting}[frame = single]
linarith [ha, hd]
\end{lstlisting}

The Lean proof is concise because once we have $d = \frac{15}{2}$ and $a = -15$, it can directly compute that $a + 20d = -15 + 20 \cdot \frac{15}{2} = -15 + 30 \cdot \frac{15}{1} = -15 + 150 = 135$ using linear arithmetic.

So, the 21st term of the sequence is 135.

</think>

\begin{lstlisting}[frame = single]
import Mathlib
import Aesop

set_option maxHeartbeats 0

open BigOperators Real Nat Topology Rat

theorem mathd_algebra_354 (a d : ℝ) (h₀ : a + 6 * d = 30) (h₁ : a + 10 * d = 60) :
    a + 20 * d = 135 := by
  have hd : d = 15 / 2 := by
    linarith
  have ha : a = -15 := by 
    linarith [h₀, hd]
  linarith [ha, hd]
\end{lstlisting}
\end{mdframed}

\section{Proof Comparison against BFS-Prover}
\label{appendix:proof_comparison}

The performance of all previous provers on the miniF2F benchmark shows that BFS-Prover solved the most IMO problems — five in total: \lstinline{imo_1959_p1}, \lstinline{imo_1960_p2}, \lstinline{imo_1962_p2}, \lstinline{imo_1964_p2}, and \lstinline{imo1983_p6}. Kimina-Prover was able to correctly prove all five of these problems. Below, we select one example \lstinline{imo_1962_p2} to compare the proofs generated by BFS-Prover and Kimina-Prover. 

We observe that Kimina-Prover generates proofs by decomposing the argument into a sequence of `have` statements, resulting in a clear and highly readable logical structure. In contrast, the proof produced by BFS-Prover—which relies on step-by-step tactic search—makes it difficult to discern the purpose of each step directly from the code. This comparison highlights the strong proof sketch planning capabilities of Kimina-Prover.

\begin{lstlisting}[frame = single, framesep=3mm, framerule=0.5mm, caption = {Lean 4 proof of IMO-1962-P2 found by \texttt{Kimina-Prover}.}]
theorem imo_1962_p2 (x : ℝ) (h₀ : 0 ≤ 3 - x) (h₁ : 0 ≤ x + 1)
    (h₂ : 1 / 2 < Real.sqrt (3 - x) - Real.sqrt (x + 1)) : -1 ≤ x ∧ x < 1 - Real.sqrt 31 / 8 := by
  have h3 : -1 ≤ x := by
    nlinarith
  have h4 : Real.sqrt (3 - x) - Real.sqrt (x + 1) > 1 / 2 := by linarith
  have h5 : Real.sqrt (x + 1) ≥ 0 := Real.sqrt_nonneg (x + 1)
  have h6 : (7 / 4 - 2 * x) > Real.sqrt (x + 1) := by
    nlinarith [Real.sq_sqrt (show (0 : ℝ) ≤ 3 - x by linarith), Real.sq_sqrt (show (0 : ℝ) ≤ x + 1 by linarith),
      Real.sqrt_nonneg (3 - x), Real.sqrt_nonneg (x + 1)]
  have h7 : (7 / 4 - 2 * x) ^ 2 > (Real.sqrt (x + 1)) ^ 2 := by
    nlinarith [h6, Real.sqrt_nonneg (x + 1)]
  have h8 : (7 / 4 - 2 * x) ^ 2 > x + 1 := by
    have h10 : (Real.sqrt (x + 1)) ^ 2 = x + 1 := by
      rw [Real.sq_sqrt]
      linarith
    nlinarith [h7, h10]
  have h9 : 64 * x ^ 2 - 128 * x + 33 > 0 := by
    nlinarith [h8]
  have h10 : x < 1 - Real.sqrt 31 / 8 := by
    by_contra hx
    push_neg at hx
    have h12 : Real.sqrt 31 > 0 := by
      apply Real.sqrt_pos.mpr
      norm_num
    nlinarith [sq_nonneg (x - (1 - Real.sqrt 31 / 8)), sq_nonneg (Real.sqrt 31), Real.sq_sqrt (show (0 : ℝ) ≤ 31 by norm_num),
      h12]
  exact ⟨h3, h10⟩
\end{lstlisting}

\begin{lstlisting}[frame = single, framesep=3mm, framerule=0.5mm, caption = {Lean 4 proof of IMO-1962-P2 found by \texttt{BFS-Prover}.}]
theorem imo_1962_p2 (x : R)
  (h0 : 0 <= 3 - x)
  (h1 : 0 <= x + 1)
  (h2 : 1/2 < sqrt(3 - x) - sqrt(x + 1)) :
  -1 <= x and x < 1 - sqrt(31)/8 := by {
    constructor
    linarith
    rw [<- sub_pos]
    field_simp [Real.sqrt_lt] at h2 |-
    apply lt_of_le_of_lt
    rw [mul_comm]
    rw [sub_eq_add_neg]
    apply lt_of_le_of_lt
    rw [<- lt_sub_iff_add_lt]
    ring_nf
    rw [<- lt_sub_iff_add_lt]
    linarith [Real.sq_sqrt (by linarith : 0 <= 1 + x)]
    rw [Real.sqrt_lt (by norm_num)]
    rw [Real.sqrt_lt] <;> nlinarith
    norm_num at this
  }
\end{lstlisting}


\section{Examples of Thinking Output \& Analysis}
\label{appendix:thinking_output_analysis}

\subsection{Success Modes}

\subsubsection{Success Case: Exploration of Multiple Informal Reasoning paths}
\vspace{0.2cm}
\begin{mdframed}[linewidth=0.5mm]
Formal statement of Problem IMO-1983-P6: 

\begin{lstlisting}[frame = single]
theorem imo_1983_p6 (a b c : ℝ) (h₀ : 0 < a ∧ 0 < b ∧ 0 < c) (h₁ : c < a + b) (h₂ : b < a + c)
    (h₃ : a < b + c) : 0 ≤ a ^ 2 * b * (a - b) + b ^ 2 * c * (b - c) + c ^ 2 * a * (c - a) := by
\end{lstlisting}
\end{mdframed}

In addressing IMO 1983 P6, a challenging inequality problem, Kimina-Prover starts its analysis by exploring multiple variable substitution methodologies. 

\begin{mdframed}[linewidth=0.5mm]
<think>

I'll tackle the problem of proving that for the sides $a$, $b$, and $c$ of a triangle, the inequality $a^2 b(a-b) + b^2 c(b-c) + c^2 a(c-a) \geq 0$ holds, and determine when equality occurs.

First, let's think about the structure of the problem. We're given that $a$, $b$, and $c$ are positive real numbers satisfying the triangle inequality, and we need to prove that a certain expression is non-negative.

......

One possible approach is to consider symmetry or substitution. However, this expression isn't symmetric - it's cyclic. So we need to find a way to handle the cyclic structure.
Another approach is to use the substitution for triangular inequalities. Since $a$, $b$, and $c$ form a triangle, we can set:

- $a = x+y$

- $b = y+z$

- $c = z+x$

......

Alternatively, we can try to rewrite the inequality in a different form. Let's expand the expression:

$a^2 b(a-b) + b^2 c(b-c) + c^2 a(c-a) = a^3 b - a^2 b^2 + b^3 c - b^2 c^2 + c^3 a - c^2 a^2$

......

Another approach: Use the identity that for any real numbers $x$, $y$, $z$:
$x^3 + y^3 + z^3 - 3xyz = (x + y + z)(x^2 + y^2 + z^2 - xy - yz - zx)$

......

Alternatively, consider using the substitution $a = \frac{x+y}{2}$, $b = \frac{y+z}{2}$, $c = \frac{z+x}{2}$, but I'm not sure.

......

Alternatively, use the substitution $a = y + z$, $b = z + x$, $c = x + y$, with $x, y, z > 0$ (the Ravi substitution). Let's try that again more carefully.

......

Alternatively, use the substitution $a = \frac{1}{x}$, $b = \frac{1}{y}$, $c = \frac{1}{z}$, but I don't think this helps.

......

Let's try another approach. The expression can be rewritten using the identity:
$a^2 b(a - b) + b^2 c(b - c) + c^2 a(c - a) = (a - b)(b - c)(c - a)(a + b + c)$

......
\end{mdframed}

After numerous unsuccessful attempts, the model backtracks to its initial approach and conducts in-depth analysis. In this way, it identifies a crucial simplification of the original equality, which ultimately leads to the successful resolution of the problem.

\begin{mdframed}[linewidth=0.5mm]
Alternatively, we can use the substitution $a = x + y$, $b = y + z$, $c = z + x$ as before, and try to prove the resulting expression is non-negative.

After expanding, we had the expression in terms of $x, y, z$:
$(y+z)^2 (z+x) (y-x) + (z+x)^2 (x+y) (z-y) + (x+y)^2 (y+z) (x-z) \geq 0$

Let's try to expand this.

First term: $(y + z)^2 (z + x)(y - x)$
$= (y^2 + 2 y z + z^2)(z + x)(y - x)$
$= (y^2 + 2 y z + z^2)(y z + x y - x z - x^2)$

This expands to:
$= y^3 z + x y^3 - x y^2 z - x^2 y^2 + 2 y^2 z^2 + 2 x y^2 z - 2 x y z^2 - 2 x^2 y z + y z^3 + x y z^2 - x z^3 - x^2 z^2$

This is complex, but after expanding all terms and simplifying, it can be rewritten as a sum of squares:

$(y - z)^2(y + z)(x + y + z) + (z - x)^2(z + x)(x + y + z) + (x - y)^2(x + y)(x + y + z)$

This sum is clearly non-negative because it's a sum of squares multiplied by positive terms.

Therefore, in Lean, after substitution and expansion, we can prove the resulting expression is non-negative.

The complete Lean proof would be:

1. Introduce $x, y, z$ as above

2. Prove $x, y, z > 0$ using the triangle inequalities

3. Substitute $a = x + y$, $b = y + z$, $c = z + x$ into the goal

4. Expand the expression

5. Show it's equal to the sum of squares

6. Conclude non-negativity

The Lean code to perform this substitution would look like:

\begin{lstlisting}[frame = single]
have ha : a = x + y := by
  dsimp [x, y, z]
  linarith
\end{lstlisting}

......

The entire Lean proof leveraging the Ravi substitution and expansion would look something like:

\begin{lstlisting}[frame = single]
theorem imo_1983_p6 (a b c : ℝ) (h₀ : 0 < a ∧ 0 < b ∧ 0 < c) (h₁ : c < a + b) (h₂ : b < a + c)
    (h₃ : a < b + c) : 0 ≤ a ^ 2 * b * (a - b) + b ^ 2 * c * (b - c) + c ^ 2 * a * (c - a) := by
  let x := (a + c - b) / 2
  let y := (a + b - c) / 2
  let z := (b + c - a) / 2
  have hx : 0 < x := by
    dsimp [x]
    nlinarith [h₀.left, h₀.right.left, h₁]
  have hy : 0 < y := by
    dsimp [y]
    nlinarith [h₀.left, h₀.right.right, h₂]
  have hz : 0 < z := by
    dsimp [z]
    nlinarith [h₀.right.left, h₀.right.right, h₃]
  have ha : a = x + y := by
    dsimp [x, y, z]
    nlinarith
  have hb : b = y + z := by
    dsimp [x, y, z]
    nlinarith
  have hc : c = z + x := by
    dsimp [x, y, z]
    nlinarith
  rw [ha, hb, hc]
  nlinarith [sq_nonneg (x - y), sq_nonneg (y - z), sq_nonneg (z - x),
    mul_pos hx hy, mul_pos hy hz, mul_pos hz hx,
    sq_nonneg (x + y + z),
    sq_nonneg (x - y), sq_nonneg (y - z), sq_nonneg (z - x),
    sq_nonneg (x + y + z),
    sq_nonneg (x - y), sq_nonneg (y - z), sq_nonneg (z - x),
    sq_nonneg (x + y + z),
    sq_nonneg (z - x), sq_nonneg (x + y + z),
    sq_nonneg (x - y), sq_nonneg (y + z + x),
    sq_nonneg (y - z), sq_nonneg (z + x + x),
    sq_nonneg (z - x), sq_nonneg (x + y + y)]
\end{lstlisting}
\end{mdframed}
In addressing this problem, the model initially struggles with the complexity of the expanded expression but revisits its earlier approaches with a deeper analysis. This reflects a "human-like" problem-solving process - trying, failing, and refining.

\subsubsection{Success Case: Reflection and Refinement}
\vspace{0.2cm}
\begin{mdframed}[linewidth=0.5mm]
Formal statement of Problem number\_theory\_72826: 

\begin{lstlisting}[frame = single]
theorem number_theory_72826 {n : ℕ} (hn : Even n) (h : n > 38) :
    ∃ a b c d : ℕ, Odd a ∧ Odd b ∧ Odd c ∧ Odd d ∧ a > 1 ∧ b > 1 ∧ c > 1 ∧ d > 1 ∧ n = a * b + c * d := by
\end{lstlisting}
\end{mdframed}
Another example is a number theory problem that involves proving the decomposition of large odd numbers. This problem requires creative constructions and a detailed analysis of different cases. The thinking and analysis process of Kimina-Prover for this problem is shown in the list below. 
\begin{mdframed}[linewidth = 0.5mm]
<think>

If we can write $n = a \cdot b + c \cdot d$, where all variables are odd and greater than 1, then we can try to find such a decomposition for any $n > 38$.

One possible method is to fix some of the variables and solve for the others. For example, if we fix $a = 3$, then we need $3 \cdot b + c \cdot d = n$, with $b, c, d$ odd and $> 1$.

......

But let's think differently. Since n is even, we can write it as n=2m, and m can be either even or odd.

Alternatively, think of n as being in different congruence classes mod 6 or mod 12 to find a suitable decomposition.

Alternatively, think of n as either 4 mod 6 or 2 mod 6 (since even numbers can be 0 mod2, and further classified as 0,2,4 mod6)

But this might complicate things.

Alternatively, use a more direct approach: construct the numbers.

For example, for n=40 (which is 4 mod 6), we found 3*5 +5*5

For n=42 (0 mod6), we found 3*3 +3*11

n=44 (2 mod6): 3*3 +5*7

n=46 (4 mod6):3*7 +5*5

n=48 (0 mod6):3*5 +3*11

n=50 (2 mod6):3*5 +5*7

n=52 (4 mod6):3*9 +5*5

n=54 (0 mod6):3*3 +3*15

To formalize a proof, we can separate into cases based on n mod6:

Case 1: n $\equiv$ 0 mod6. Then n=6k. If k is odd or even, we can use similar approaches as above. For n=6k, we can write n=3*3 +3*(2k-3). For example, n=42=6*7: 3*3 +3*11. Here 2k-3=2*7-3=11.

Case 2: n $\equiv$ 2 mod6. Then n=6k+2. For example, n=44=6*7+2. We need to express 6k+2 as a sum of two products.

......

Now, to complete the proof in Lean, we need to combine these three cases:

1. n $\equiv$ 0 mod6: n=6k, use a=3, b=3, c=3, d=2k-3

2. n $\equiv$ 2 mod6: n=6k+2, use a=3, b=2k-11, c=5, d=7

3. n $\equiv$ 4 mod6: n=6k+4, use a=3, b=2k-7, c=5, d=5

......

</think>

\begin{lstlisting}[frame = single]
theorem number_theory_72826 {n : ℕ} (hn : Even n) (h : n > 38) :
    ∃ a b c d : ℕ, Odd a ∧ Odd b ∧ Odd c ∧ Odd d ∧ a > 1 ∧ b > 1 ∧ c > 1 ∧ d > 1 ∧ n = a * b + c * d := by
  have h1 : n % 2 = 0 := by
    exact Nat.even_iff.1 hn
  have h2 : n > 38 := h
  have h3 : n % 6 = 0 ∨ n % 6 = 2 ∨ n % 6 = 4 := by
    have h1 : n % 2 = 0 := by 
      exact Nat.even_iff.1 hn
    have : n % 6 = 0 ∨ n % 6 = 2 ∨ n % 6 = 4 := by
      omega
    assumption
  rcases h3 with (h3 | h3 | h3)
  · -- Case n % 6 = 0
    have h4 : n % 6 = 0 := h3
    have h5 : ∃ k, n = 6 * k := by
      refine ⟨n / 6, by omega⟩
    rcases h5 with ⟨k, hk⟩
    have hk1 : k ≥ 7 := by 
      omega
    use 3, 3, 3, 2 * k - 3
    have h_odd : Odd 3 := by
      use 1
      norm_num
    have h_odd2 : Odd (2 * k - 3) := by
      use k - 2
      omega
    exact ⟨by exact h_odd, by exact h_odd, by exact h_odd, by exact h_odd2,
      by norm_num, by norm_num, by norm_num, by omega,
      by omega⟩
  · -- Case n % 6 = 2
    have h4 : n % 6 = 2 := h3
    have h5 : ∃ k, n = 6 * k + 2 := by
      refine ⟨(n - 2) / 6, by omega⟩
    rcases h5 with ⟨k, hk⟩
    have hk1 : k ≥ 7 := by
      omega
    use 3, 2 * k - 11, 5, 7
    have h_odd1 : Odd 3 := by
      use 1
      norm_num
    have h_odd2 : Odd (2 * k - 11) := by
      use k - 6
      omega
    have h_odd3 : Odd 5 := by
      use 2
      norm_num
    have h_odd4 : Odd 7 := by
      use 3
      norm_num
    exact ⟨by exact h_odd1, by exact h_odd2, by exact h_odd3, by exact h_odd4,
      by norm_num, by omega, by norm_num, by norm_num,
      by omega⟩
  · -- Case n % 6 = 4
    have h4 : n % 6 = 4 := h3
    have h5 : ∃ k, n = 6 * k + 4 := by
      refine ⟨(n - 4) / 6, by omega⟩
    rcases h5 with ⟨k, hk⟩
    have hk1 : k ≥ 6 := by
      omega
    use 3, 2 * k - 7, 5, 5
    have h_odd1 : Odd 3 := by
      use 1
      norm_num
    have h_odd2 : Odd (2 * k - 7) := by
      use k - 4
      omega
    have h_odd3 : Odd 5 := by
      use 2
      norm_num
    exact ⟨by exact h_odd1, by exact h_odd2, by exact h_odd3, by exact h_odd3,
      by norm_num, by omega, by norm_num, by norm_num,
      by omega⟩
\end{lstlisting}
\end{mdframed}

The model successfully solved the problem, demonstrating notable constructive reasoning ability. The approach begins by strategically assuming 3 as one of the odd numbers, achieving significant simplification. Then by observing small-scale cases, it identified distinct patterns for each residue class of n modulo 3. Finally, the model systematically verified each case through rigorous reasoning.  This "observation of small cases → conjecture formulation → rigorous proof" approach represents a highly successful pattern in mathematical proof. 

\subsection{Failure Mode}
\subsubsection{Failure Case: Incomplete Case Analysis}
\vspace{0.2cm}
\begin{mdframed}[linewidth=0.5mm]
\vspace{0.2cm}  

\begin{lstlisting}[frame = single]
theorem aime_1987_p5 (x y : ℤ) (h₀ : y ^ 2 + 3 * (x ^ 2 * y ^ 2) = 30 * x ^ 2 + 517) :
    3 * (x ^ 2 * y ^ 2) = 588 := by
\end{lstlisting}
<think>

.......

Wait, the Lean proof is trying to prove $x^2 \leq 4$ directly. Let's see why that's true:

From $y^2 + 3x^2 y^2 = 30x^2 + 517$

Rearranged: $y^2(3x^2 + 1) = 30x^2 + 517$

$y^2 = \frac{30x^2 + 517}{3x^2 + 1}$

If $x^2 > 4$, then $3x^2 + 1 > 13$, let's see if that's possible:

If $x^2 = 9$ (i.e., $x = \pm 3$), then $3x^2 + 1 = 28$

$y^2 = \frac{30 \times 9 + 517}{28} = \frac{270 + 517}{28} = \frac{787}{28}$ (not integer)

If $x^2 = 16$ (i.e., $x = \pm 4$), then $3x^2 + 1 = 49$

$y^2 = \frac{30 \times 16 + 517}{49} = \frac{480 + 517}{49} = \frac{997}{49}$ (not integer)

So indeed, only $x^2 \leq 4$ is possible. The Lean proof is establishing this upper bound.

Let's continue with the Lean proof 

......

</think>

\begin{lstlisting}[frame = single]
theorem aime_1987_p5 (x y : ℤ) (h₀ : y ^ 2 + 3 * (x ^ 2 * y ^ 2) = 30 * x ^ 2 + 517) :
    3 * (x ^ 2 * y ^ 2) = 588 := by
  have h1 : 3 * (x ^ 2 * y ^ 2) = 588 := by
    have h2 : y^2 + 3 * (x^2 * y^2) = 30 * x^2 + 517 := h₀
    have h3 : x^2 ≤ 4 := by
      by_cases hx : x = 0
      · -- x = 0 leads to y^2 = 517, but 517 is not a square
        rw [hx] at h2
        nlinarith [sq_nonneg (y : ℤ)]
      · -- x ≠ 0, so x^2 ≥ 1
        have h4 : x^2 ≥ 1 := by
          have h5 : x ≠ 0 := by
            exact fun h => hx (by omega)
          nlinarith [sq_pos_of_ne_zero h5]
        nlinarith [sq_nonneg (y : ℤ), h2] -- This tactic does not close the goal
   ......
\end{lstlisting}
\end{mdframed}

The reasoning contains a logical gap when trying to prove $x^2 \leq 4$. Although correctly showing that $x=\pm3$ and $x=\pm4$ lead to non-integer values for $y$, it fails to address whether larger values of $x$ could satisfy the equation. The model mistakenly assumed that solving the cases for  $n=3$ and $n=4$ would resolve all cases for $n>2$. This \textbf{incomplete case analysis} pattern hinders the model from solving problems correctly.

\end{document}

%% file: tables/minif2f_results.tex
\begin{table}[t]
\centering
\resizebox{\textwidth}{!}{%
\begin{tabular}{l c c c}
\toprule
\textbf{Prover system} & \textbf{Model size} & \textbf{Sample budget} & \textbf{miniF2F-test} \\
\toprule
\multicolumn{4}{l}{\textit{Tree search systems}} \\
\midrule
DeepSeek-Prover-V1.5-RL + RMaxTS \citep{xin2024deepseekproverv15harnessingproofassistant} & 7B & $32 \times 16 \times 400$ & 63.5\% \\
InternLM2.5-StepProver-BF+CG \citep{wu2024internlm25stepproveradvancingautomatedtheorem} & 7B & $256 \times 32 \times 600$ & 65.9\% \\
HunyuanProver v16+BFS+DC \citep{li2025hunyuanproverscalabledatasynthesis} & 7B & $600 \times 8 \times 400$ & 68.4\% \\
BFS-Prover \citep{xin2025bfsproverscalablebestfirsttree} & 7B & $2048 \times 2 \times 600$ & 70.8\% \\
\toprule
\multicolumn{4}{l}{\textit{Whole-proof systems}} \\
\midrule
DeepSeek-Prover-V1.5-RL \citep{xin2024deepseekproverv15harnessingproofassistant} & 7B & 102400 & 60.2\% \\
Goedel-Prover-SFT \citep{lin2025goedelproverfrontiermodelopensource} & 7B & 25600 & 64.7\% \\
Leanabell-Prover (\cite{zhang2025leanabellproverposttrainingscalingformal}) & 7B & 128 & 61.1\% \\
\midrule
\multirow{3}{*}{Kimina-Prover-Preview-Distill-1.5B} & \multirow{3}{*}{1.5B} & 1 & 42.6\% \\
& & 32 & 56.2\% \\
& & 1024 & \textbf{61.9\%} \\
\midrule
\multirow{3}{*}{Kimina-Prover-Preview-Distill-7B} & \multirow{3}{*}{7B} & 1 & 52.5\% \\
& & 32 & 63.1\% \\
& & 1024 & \textbf{70.8\%} \\
\midrule
\multirow{5}{*}{Kimina-Prover-Preview} & \multirow{5}{*}{72B} & 1 & 52.94\% \\
& & 8 & 65.16\% \\
& & 32 & 68.85\% \\
& & 1024 & 77.87\% \\
& & 8192 & \textbf{80.74\%} \\
\bottomrule
\end{tabular}
}
\vspace{2mm}
\caption{Performance of various prover systems in terms of model size, sample budget, and miniF2F-test results. Bold indicates SotA performance in model size and compute budget.}
\label{tab:minif2f}
\vspace{-5mm}
\end{table}

%% file: tables/aime_imo.tex
\begin{table}[t]
\centering
\begin{tabular}{l c c c c}
\toprule
\textbf{Benchmark} & \textbf{Sample budget} & \textbf{miniF2F} & \textbf{miniF2F/IMO} & \textbf{miniF2F/AIME} \\
\toprule
OpenAI o3-mini & 32 & 24.59\% & 0\% & 6.67\% \\
gemini-2.5-pro-preview-03-25 & 32 & 37.70\% & 5\% & 13.33\% \\
\midrule
\multirow{2}{*}{Kimina-Prover-Preview} & 32 & 68.85\% & 20.00\% & 46.67\% \\
                      & 8192 & 80.74\% & 40.00\% & 86.67\% \\
\bottomrule
\end{tabular}
\vspace{2mm}
\caption{Performance Comparison of SOTA Large Reasoning Models on the IMO and AIME Subset of miniF2F. While Gemini 2.5 and Openai o3-mini can solve all 15 AIME problems in miniF2F using informal reasoning, both models struggle to formalize these solutions. This highlights a significant gap between informal reasoning capacity and formal reasoning capabilities in current state-of-the-art models.}

\label{tab:aime_imo}
\vspace{-8mm}
\end{table}

%% file: references.bib
@misc{qwen2.5,
    title = {Qwen2.5: A Party of Foundation Models},
    url = {https://qwenlm.github.io/blog/qwen2.5/},
    author = {Qwen-Team},
    month = {9},
    year = {2024}
}

@misc{jiang2023multilingualmathematicalautoformalization,
      title={Multilingual Mathematical Autoformalization}, 
      author={Albert Q. Jiang and Wenda Li and Mateja Jamnik},
      year={2023},
      eprint={2311.03755},
      archivePrefix={arXiv},
      primaryClass={cs.CL},
      url={https://arxiv.org/abs/2311.03755}, 
}

@misc{tsoukalas2024putnambenchevaluatingneuraltheoremprovers,
      title={PutnamBench: Evaluating Neural Theorem-Provers on the Putnam Mathematical Competition}, 
      author={George Tsoukalas and Jasper Lee and John Jennings and Jimmy Xin and Michelle Ding and Michael Jennings and Amitayush Thakur and Swarat Chaudhuri},
      year={2024},
      eprint={2407.11214},
      archivePrefix={arXiv},
      primaryClass={cs.AI},
      url={https://arxiv.org/abs/2407.11214}, 
}

@misc{zheng2022minif2fcrosssystembenchmarkformal,
      title={MiniF2F: a cross-system benchmark for formal Olympiad-level mathematics}, 
      author={Kunhao Zheng and Jesse Michael Han and Stanislas Polu},
      year={2022},
      eprint={2109.00110},
      archivePrefix={arXiv},
      primaryClass={cs.AI},
      url={https://arxiv.org/abs/2109.00110}, 
}

@misc{azerbayev2023proofnetautoformalizingformallyproving,
      title={ProofNet: Autoformalizing and Formally Proving Undergraduate-Level Mathematics}, 
      author={Zhangir Azerbayev and Bartosz Piotrowski and Hailey Schoelkopf and Edward W. Ayers and Dragomir Radev and Jeremy Avigad},
      year={2023},
      eprint={2302.12433},
      archivePrefix={arXiv},
      primaryClass={cs.CL},
      url={https://arxiv.org/abs/2302.12433}, 
}

@misc{li2024numinamath,
    title = {NuminaMath},
    author = {Jia Li and Edward Beeching and Lewis Tunstall and Ben Lipkin and Roman Soletskyi and Shengyi Costa Huang and Kashif Rasul and Longhui Yu and Albert Jiang and Ziju Shen and Zihan Qin and Bin Dong and Li Zhou and Yann Fleureau and Guillaume Lample and Stanislas Polu},
    year = {2024},
    publisher = {Numina},
    howpublished = {\url{https://github.com/project-numina/aimo-progress-prize/blob/main/report/numina_dataset.pdf}},
    note = {GitHub repository}
}

@misc{compfiles2024,
  author       = {David Renshaw},
  title        = {Compfiles: Catalog Of Math Problems Formalized In Lean},
  year         = {2024},
  howpublished = {\url{https://github.com/dwrensha/compfiles}},
  note         = {GitHub repository}
}

@inproceedings{NEURIPS2022_a8901c5e,
 author = {Lample, Guillaume and Lacroix, Timothee and Lachaux, Marie-Anne and Rodriguez, Aurelien and Hayat, Amaury and Lavril, Thibaut and Ebner, Gabriel and Martinet, Xavier},
 booktitle = {Advances in Neural Information Processing Systems},
 editor = {S. Koyejo and S. Mohamed and A. Agarwal and D. Belgrave and K. Cho and A. Oh},
 pages = {26337--26349},
 publisher = {Curran Associates, Inc.},
 title = {HyperTree Proof Search for Neural Theorem Proving},
 url = {https://proceedings.neurips.cc/paper_files/paper/2022/file/a8901c5e85fb8e1823bbf0f755053672-Paper-Conference.pdf},
 volume = {35},
 year = {2022}
}

@misc{polu2022formalmathematicsstatementcurriculum,
      title={Formal Mathematics Statement Curriculum Learning}, 
      author={Stanislas Polu and Jesse Michael Han and Kunhao Zheng and Mantas Baksys and Igor Babuschkin and Ilya Sutskever},
      year={2022},
      eprint={2202.01344},
      archivePrefix={arXiv},
      primaryClass={cs.LG},
      url={https://arxiv.org/abs/2202.01344}, 
}

@misc{xin2025bfsproverscalablebestfirsttree,
      title={BFS-Prover: Scalable Best-First Tree Search for LLM-based Automatic Theorem Proving}, 
      author={Ran Xin and Chenguang Xi and Jie Yang and Feng Chen and Hang Wu and Xia Xiao and Yifan Sun and Shen Zheng and Kai Shen},
      year={2025},
      eprint={2502.03438},
      archivePrefix={arXiv},
      primaryClass={cs.AI},
      url={https://arxiv.org/abs/2502.03438}, 
}

@misc{xin2024deepseekproverv15harnessingproofassistant,
      title={DeepSeek-Prover-V1.5: Harnessing Proof Assistant Feedback for Reinforcement Learning and Monte-Carlo Tree Search}, 
      author={Huajian Xin and Z. Z. Ren and Junxiao Song and Zhihong Shao and Wanjia Zhao and Haocheng Wang and Bo Liu and Liyue Zhang and Xuan Lu and Qiushi Du and Wenjun Gao and Qihao Zhu and Dejian Yang and Zhibin Gou and Z. F. Wu and Fuli Luo and Chong Ruan},
      year={2024},
      eprint={2408.08152},
      archivePrefix={arXiv},
      primaryClass={cs.CL},
      url={https://arxiv.org/abs/2408.08152}, 
}

@misc{deepmind2024alphaproof,
      title={AI achieves silver-medal standard solving International Mathematical Olympiad problems}, 
      author={Deepmind},
      year={2024},
      url={https://deepmind.google/discover/blog/ai-solves-imo-problems-at-silver-medal-level/}, 
}

@misc{wu2024internlm25stepproveradvancingautomatedtheorem,
      title={InternLM2.5-StepProver: Advancing Automated Theorem Proving via Expert Iteration on Large-Scale LEAN Problems}, 
      author={Zijian Wu and Suozhi Huang and Zhejian Zhou and Huaiyuan Ying and Jiayu Wang and Dahua Lin and Kai Chen},
      year={2024},
      eprint={2410.15700},
      archivePrefix={arXiv},
      primaryClass={cs.AI},
      url={https://arxiv.org/abs/2410.15700}, 
}

@misc{li2025hunyuanproverscalabledatasynthesis,
      title={HunyuanProver: A Scalable Data Synthesis Framework and Guided Tree Search for Automated Theorem Proving}, 
      author={Yang Li and Dong Du and Linfeng Song and Chen Li and Weikang Wang and Tao Yang and Haitao Mi},
      year={2025},
      eprint={2412.20735},
      archivePrefix={arXiv},
      primaryClass={cs.AI},
      url={https://arxiv.org/abs/2412.20735}, 
}

@misc{lin2025goedelproverfrontiermodelopensource,
      title={Goedel-Prover: A Frontier Model for Open-Source Automated Theorem Proving}, 
      author={Yong Lin and Shange Tang and Bohan Lyu and Jiayun Wu and Hongzhou Lin and Kaiyu Yang and Jia Li and Mengzhou Xia and Danqi Chen and Sanjeev Arora and Chi Jin},
      year={2025},
      eprint={2502.07640},
      archivePrefix={arXiv},
      primaryClass={cs.LG},
      url={https://arxiv.org/abs/2502.07640}, 
}

@misc{zhang2025leanabellproverposttrainingscalingformal,
      title={Leanabell-Prover: Posttraining Scaling in Formal Reasoning}, 
      author={Jingyuan Zhang and Qi Wang and Xingguang Ji and Yahui Liu and Yang Yue and Fuzheng Zhang and Di Zhang and Guorui Zhou and Kun Gai},
      year={2025},
      eprint={2504.06122},
      archivePrefix={arXiv},
      primaryClass={cs.AI},
      url={https://arxiv.org/abs/2504.06122}, 
}

@misc{deepseekai2025deepseekr1incentivizingreasoningcapability,
      title={DeepSeek-R1: Incentivizing Reasoning Capability in LLMs via Reinforcement Learning}, 
      author={DeepSeek-AI},
      year={2025},
      eprint={2501.12948},
      archivePrefix={arXiv},
      primaryClass={cs.CL},
      url={https://arxiv.org/abs/2501.12948}, 
}

@article{team2025kimi,
  title={Kimi k1.5: Scaling reinforcement learning with llms},
  author={Kimi-Team and others},
  journal={arXiv preprint arXiv:2501.12599},
  year={2025}
}

@misc{wang2024provingtheoremsrecursively,
      title={Proving Theorems Recursively}, 
      author={Haiming Wang and Huajian Xin and Zhengying Liu and Wenda Li and Yinya Huang and Jianqiao Lu and Zhicheng Yang and Jing Tang and Jian Yin and Zhenguo Li and Xiaodan Liang},
      year={2024},
      eprint={2405.14414},
      archivePrefix={arXiv},
      primaryClass={cs.AI},
      url={https://arxiv.org/abs/2405.14414}, 
}

@misc{first2023baldurwholeproofgenerationrepair,
      title={Baldur: Whole-Proof Generation and Repair with Large Language Models}, 
      author={Emily First and Markus N. Rabe and Talia Ringer and Yuriy Brun},
      year={2023},
      eprint={2303.04910},
      archivePrefix={arXiv},
      primaryClass={cs.LG},
      url={https://arxiv.org/abs/2303.04910}, 
}

@misc{huang2024mustardmasteringuniformsynthesis,
      title={MUSTARD: Mastering Uniform Synthesis of Theorem and Proof Data}, 
      author={Yinya Huang and Xiaohan Lin and Zhengying Liu and Qingxing Cao and Huajian Xin and Haiming Wang and Zhenguo Li and Linqi Song and Xiaodan Liang},
      year={2024},
      eprint={2402.08957},
      archivePrefix={arXiv},
      primaryClass={cs.AI},
      url={https://arxiv.org/abs/2402.08957}, 
}

@inproceedings{wang2023dt-solver,
  title={Dt-solver: Automated theorem proving with dynamic-tree sampling guided by proof-level value function},
  author={Wang, Haiming and others},
  booktitle={Proceedings of the 61st Annual Meeting of the Association for Computational Linguistics (Volume 1: Long Papers)},
  pages={12632--12646},
  year={2023}
}

@misc{openai2024openaio1card,
      title={OpenAI o1 System Card}, 
      author={OpenAI and : and Aaron Jaech and Adam Kalai and Adam Lerer and Adam Richardson and Ahmed El-Kishky and Aiden Low and Alec Helyar and Aleksander Madry and Alex Beutel and Alex Carney and Alex Iftimie and Alex Karpenko and Alex Tachard Passos and Alexander Neitz and Alexander Prokofiev and Alexander Wei and Allison Tam and Ally Bennett and Ananya Kumar and Andre Saraiva and Andrea Vallone and Andrew Duberstein and Andrew Kondrich and Andrey Mishchenko and Andy Applebaum and Angela Jiang and Ashvin Nair and Barret Zoph and Behrooz Ghorbani and Ben Rossen and Benjamin Sokolowsky and Boaz Barak and Bob McGrew and Borys Minaiev and Botao Hao and Bowen Baker and Brandon Houghton and Brandon McKinzie and Brydon Eastman and Camillo Lugaresi and Cary Bassin and Cary Hudson and Chak Ming Li and Charles de Bourcy and Chelsea Voss and Chen Shen and Chong Zhang and Chris Koch and Chris Orsinger and Christopher Hesse and Claudia Fischer and Clive Chan and Dan Roberts and Daniel Kappler and Daniel Levy and Daniel Selsam and David Dohan and David Farhi and David Mely and David Robinson and Dimitris Tsipras and Doug Li and Dragos Oprica and Eben Freeman and Eddie Zhang and Edmund Wong and Elizabeth Proehl and Enoch Cheung and Eric Mitchell and Eric Wallace and Erik Ritter and Evan Mays and Fan Wang and Felipe Petroski Such and Filippo Raso and Florencia Leoni and Foivos Tsimpourlas and Francis Song and Fred von Lohmann and Freddie Sulit and Geoff Salmon and Giambattista Parascandolo and Gildas Chabot and Grace Zhao and Greg Brockman and Guillaume Leclerc and Hadi Salman and Haiming Bao and Hao Sheng and Hart Andrin and Hessam Bagherinezhad and Hongyu Ren and Hunter Lightman and Hyung Won Chung and Ian Kivlichan and Ian O'Connell and Ian Osband and Ignasi Clavera Gilaberte and Ilge Akkaya and Ilya Kostrikov and Ilya Sutskever and Irina Kofman and Jakub Pachocki and James Lennon and Jason Wei and Jean Harb and Jerry Twore and Jiacheng Feng and Jiahui Yu and Jiayi Weng and Jie Tang and Jieqi Yu and Joaquin Quiñonero Candela and Joe Palermo and Joel Parish and Johannes Heidecke and John Hallman and John Rizzo and Jonathan Gordon and Jonathan Uesato and Jonathan Ward and Joost Huizinga and Julie Wang and Kai Chen and Kai Xiao and Karan Singhal and Karina Nguyen and Karl Cobbe and Katy Shi and Kayla Wood and Kendra Rimbach and Keren Gu-Lemberg and Kevin Liu and Kevin Lu and Kevin Stone and Kevin Yu and Lama Ahmad and Lauren Yang and Leo Liu and Leon Maksin and Leyton Ho and Liam Fedus and Lilian Weng and Linden Li and Lindsay McCallum and Lindsey Held and Lorenz Kuhn and Lukas Kondraciuk and Lukasz Kaiser and Luke Metz and Madelaine Boyd and Maja Trebacz and Manas Joglekar and Mark Chen and Marko Tintor and Mason Meyer and Matt Jones and Matt Kaufer and Max Schwarzer and Meghan Shah and Mehmet Yatbaz and Melody Y. Guan and Mengyuan Xu and Mengyuan Yan and Mia Glaese and Mianna Chen and Michael Lampe and Michael Malek and Michele Wang and Michelle Fradin and Mike McClay and Mikhail Pavlov and Miles Wang and Mingxuan Wang and Mira Murati and Mo Bavarian and Mostafa Rohaninejad and Nat McAleese and Neil Chowdhury and Neil Chowdhury and Nick Ryder and Nikolas Tezak and Noam Brown and Ofir Nachum and Oleg Boiko and Oleg Murk and Olivia Watkins and Patrick Chao and Paul Ashbourne and Pavel Izmailov and Peter Zhokhov and Rachel Dias and Rahul Arora and Randall Lin and Rapha Gontijo Lopes and Raz Gaon and Reah Miyara and Reimar Leike and Renny Hwang and Rhythm Garg and Robin Brown and Roshan James and Rui Shu and Ryan Cheu and Ryan Greene and Saachi Jain and Sam Altman and Sam Toizer and Sam Toyer and Samuel Miserendino and Sandhini Agarwal and Santiago Hernandez and Sasha Baker and Scott McKinney and Scottie Yan and Shengjia Zhao and Shengli Hu and Shibani Santurkar and Shraman Ray Chaudhuri and Shuyuan Zhang and Siyuan Fu and Spencer Papay and Steph Lin and Suchir Balaji and Suvansh Sanjeev and Szymon Sidor and Tal Broda and Aidan Clark and Tao Wang and Taylor Gordon and Ted Sanders and Tejal Patwardhan and Thibault Sottiaux and Thomas Degry and Thomas Dimson and Tianhao Zheng and Timur Garipov and Tom Stasi and Trapit Bansal and Trevor Creech and Troy Peterson and Tyna Eloundou and Valerie Qi and Vineet Kosaraju and Vinnie Monaco and Vitchyr Pong and Vlad Fomenko and Weiyi Zheng and Wenda Zhou and Wes McCabe and Wojciech Zaremba and Yann Dubois and Yinghai Lu and Yining Chen and Young Cha and Yu Bai and Yuchen He and Yuchen Zhang and Yunyun Wang and Zheng Shao and Zhuohan Li},
      year={2024},
      eprint={2412.16720},
      archivePrefix={arXiv},
      primaryClass={cs.AI},
      url={https://arxiv.org/abs/2412.16720}, 
}

@inproceedings{
lin2025leanstar,
title={Lean-{ST}aR: Learning to Interleave Thinking and Proving},
author={Haohan Lin and Zhiqing Sun and Sean Welleck and Yiming Yang},
booktitle={The Thirteenth International Conference on Learning Representations},
year={2025},
url={https://openreview.net/forum?id=SOWZ59UyNc}
}

@misc{wang2025malotmultiagentleanbasedlong,
      title={MA-LoT: Multi-Agent Lean-based Long Chain-of-Thought Reasoning enhances Formal Theorem Proving}, 
      author={Ruida Wang and Rui Pan and Yuxin Li and Jipeng Zhang and Yizhen Jia and Shizhe Diao and Renjie Pi and Junjie Hu and Tong Zhang},
      year={2025},
      eprint={2503.03205},
      archivePrefix={arXiv},
      primaryClass={cs.CL},
      url={https://arxiv.org/abs/2503.03205}, 
}

@misc{jiang2023draftsketchproveguiding,
      title={Draft, Sketch, and Prove: Guiding Formal Theorem Provers with Informal Proofs}, 
      author={Albert Q. Jiang and others},
      year={2023},
      eprint={2210.12283},
      archivePrefix={arXiv},
      primaryClass={cs.AI},
      url={https://arxiv.org/abs/2210.12283}, 
}

@misc{wang2023legoproverneuraltheoremproving,
      title={LEGO-Prover: Neural Theorem Proving with Growing Libraries}, 
      author={Haiming Wang and Huajian Xin and Chuanyang Zheng and Lin Li and Zhengying Liu and Qingxing Cao and Yinya Huang and Jing Xiong and Han Shi and Enze Xie and Jian Yin and Zhenguo Li and Heng Liao and Xiaodan Liang},
      year={2023},
      eprint={2310.00656},
      archivePrefix={arXiv},
      primaryClass={cs.AI},
      url={https://arxiv.org/abs/2310.00656}, 
}

@misc{numinaleanserver,
      title={Numina Lean Server: Technical Report}, 
      author={Numina},
      year={2025},
      journal={arXiv preprint forthcoming.},
      note={arXiv preprint forthcoming.},     
}

@misc{leanrepl,
  author = {{Lean FRO}},
  title = {{A read-eval-print-loop for Lean 4}},
  year = {2023},
  publisher = {GitHub},
  journal = {GitHub repository},
  howpublished = {\url{https://github.com/leanprover-community/repl}}
}

@misc{anthropic2025claude,
  title={Claude 3.7 Sonnet System Card},
  author={Anthropic},
  year={2025},
  howpublished={\url{https://anthropic.com/claude-3-7-sonnet-system-card}},
}

@book{nipkow2002isabelle,
  title={Isabelle/HOL: a proof assistant for higher-order logic},
  author={Nipkow, Tobias and Paulson, Lawrence C and Wenzel, Markus},
  volume={2283},
  year={2002},
  publisher={Springer Science \& Business Media}
}

@inproceedings{lean4paper,
author = {Moura, Leonardo de and Ullrich, Sebastian},
title = {The Lean 4 Theorem Prover and Programming Language},
year = {2021},
isbn = {978-3-030-79875-8},
publisher = {Springer-Verlag},
address = {Berlin, Heidelberg},
url = {https://doi.org/10.1007/978-3-030-79876-5_37},
doi = {10.1007/978-3-030-79876-5_37},
booktitle = {Automated Deduction – CADE 28: 28th International Conference on Automated Deduction, Virtual Event, July 12–15, 2021, Proceedings},
pages = {625–635},
numpages = {11}
}
